\documentclass[twoside,11pt]{article}

\usepackage{blindtext}

\usepackage[preprint, abbrvbib]{dmlr2e}
\usepackage{multirow}
\usepackage{makecell}
\usepackage[table]{xcolor}
\usepackage{booktabs}
\usepackage{tabularx}
\usepackage{enumitem}

\newif\ifworkinprogress
\workinprogresstrue 

\ifworkinprogress

\else

\fi

\usepackage{lastpage}
\dmlrheading{23}{2026}{1-\pageref{LastPage}}{9/23; Revised 10/23}{11/23}{21-0000}{Cissée et al.} 
\def\openreview{\url{https://openreview.net/forum?id=XXXX}} 

\ShortHeadings{Hidden Data Challenges in Clinical Machine Learning}{Cissée et al.}
\firstpageno{1}

\begin{document}

\title{Hidden Data Challenges in Clinical Machine Learning: Revisiting
MIMIC Dataset}

\author{\name Carolin Cissée$^{1}$ \email Cissee.Carolin@mh-hannover.de
       \AND
       \name Dennis Fast$^{2}$
       \AND
       \name Raneen Younis$^{1,3}$
       \AND
       \name Alexander Löser$^{2}$
       \AND
       \name Wolfgang Nejdl$^{3,4}$ 
       \AND
       \name Zahra Ahmadi$^{1,3}$\email Ahmadi.Zahra@mh-hannover.de \\
       \addr$^1$ Peter L. Reichertz Institute for Medical Informatics of TU Braunschweig and Hannover Medical School, Hannover Medical School, Germany\\
       \addr$^2$ DATEXIS, Berliner Hochschule für Technik, Germany\\
       \addr$^3$ Lower Saxony Center for AI and Causal Methods in Medicine (CAIMed), Hannover, Germany\\
       \addr$^4$ L3S Research Center, Leibniz University Hannover, Germany}

\editor{My editor}

\maketitle

\begin{abstract}

Widely reused preprocessing pipelines for electronic health records can encode assumptions about when information became available, what clinical episode a label represents, and what missingness means. These assumptions are rarely audited systematically and may propagate across subsequent studies when preprocessing pipelines are reused or adapted. Using MIMIC-IV as a case study, we identify and quantify violations of three complementary forms of preprocessing alignment, i.e., temporal, semantic, and population alignment, which manifest as temporal leakage, feature-task misalignment, and biased imputation. We show how ambiguities in timestamp semantics, episode definitions, endpoint construction, modality selection, and missingness handling can alter the information available to a model and affect the interpretation of its predictions. Through analyses of MIMIC-IV data and empirical case studies, we demonstrate that such choices may remain largely invisible in aggregate predictive performance while changing the signals on which models rely. We derive practical recommendations for temporally valid, clinically aligned, and reproducible preprocessing of MIMIC-IV and related electronic health record datasets. The code for our experiments, including our proposed preprocessing, is available at: \url{https://github.com/m2l-lab/mimic-iv-preprocess}.
\end{abstract}


\section{Introduction}
Machine learning has significant potential to advance medical research and clinical decision-making, with applications spanning diagnostics, treatment personalization, and prognostic modeling (e.g., \cite{Di2023Dementia, Jabbour2022RespFailure, kodali2023multimodalAICancer, Cirillo2019PersonalizedMedicine, Zhang2026Postpartum}). Nevertheless, medical data present distinct challenges. Routine clinical data are often affected by inconsistent documentation practices, changes in available features over time, high rates of missingness, redundant entries, and documentation errors, among other issues. These issues arise in part because such data are frequently extracted from hospital information systems originally designed for billing and operational workflows rather than research. Moreover, the sensitive nature of medical data and strict privacy regulations constrain data-sharing. As a result, publicly available medical datasets play an indispensable role in enabling reproducible, comparable, and accelerated research, leading to the development of numerous benchmarks based on such data. 
Despite their importance, even widely used benchmarks are susceptible to preprocessing pitfalls. Importantly, since these preprocessing pipelines are frequently reused without sufficient evaluation, errors and questionable assumptions can be propagated across studies and become embedded in the scientific literature. In this work, we use the widely used MIMIC-IV dataset ~\citep{Johnson2023MIMIC}, which contains de-identified electronic health record (EHR) data, to highlight preprocessing pitfalls that are frequently overlooked and to discuss strategies for mitigating or eliminating them.

Previous research has identified several important sources of error in MIMIC and EHR datasets more broadly. \cite{Boag2022EHR} cataloged issues in MIMIC-III, including heart rate anomalies caused by the inclusion of neonates, artifacts such as duplicate notes, and inconsistencies in documentation resulting from nurse shift changes or departmental workflow differences, such as discharge timestamps that conflict with ICU stay records. They also emphasized the risks of assuming that data are missing at random, given that clinical judgment influences which measurements are taken and when, as well as the potential for model bias and the underrepresentation of certain groups due to issues with out-of-hospital mortality data. Although these concerns remain valid, their analysis was based solely on MIMIC-III. Certain issues, such as the inclusion of neonatal data, have since been addressed in the current version, MIMIC-IV.
\cite{Ghassemi2020EHR} provide a broader analysis of challenges in EHR data, emphasizing that the absence of certain data values may reflect a patient's condition, clinician assessments, or underlying societal biases. They also caution against information leakage between prediction tasks and input features that may occur if all available data is used without careful validation. Similarly, \cite{khaled2025review} highlight technical challenges such as heterogeneity in data granularity, high-cardinality features, and limited label precision in the MIMIC database. \cite{Sauer2022EHRPitfalls} highlight the need to incorporate knowledge from relevant healthcare providers to verify preprocessing pipelines and study design for EHR data. They also warn of potential sample selection bias due to exclusion and inclusion criteria, information leakage between the train and test sets, and differences in inter- and intraphysician decision-making affecting causal inference. They also discuss the need for precise variable and endpoint definitions that consider clinical context, as well as deployment and generalization issues, when training on data collected within the workflow and cohort of one hospital. \cite{Kapoor2023Leakage} address how leakage and irreproducible results can arise generally in scientific research, exploring how the absence of a methodologically clear separation between training and test data leads to information from the test set leaking into the training data. 

While these studies have substantially improved awareness of preprocessing challenges in general and within MIMIC pipelines specifically, several important concerns remain unaddressed, impacting the validity of methodologies, results, and conclusions in subsequent research. In this work, we extend this line of research by investigating failures of temporal, semantic, and population alignment that persist in current MIMIC-IV pipelines and manifest as temporal leakage, feature-task misalignment, selection bias, or unintended noise injection. MIMIC provides an instructive case study because it illustrates the dual-edged nature of large-scale EHR datasets. Its extensive scale and multimodal structure, including time series measurements, tabular information on patient stays, free-text notes from the MIMIC-IV-Note module ~\citep{PhysioNet-mimic-iv-note-2.2}, and imaging within the MIMIC-CXR module ~\citep{Johnson2019MIMICCXR}, make it a valuable resource for diverse medical machine learning applications. At the same time, this scale amplifies the complexity of dataset preprocessing, permitting subtle errors or mistaken assumptions to remain hidden among the multitude of decisions and steps required during pipeline implementation. 
Specifically, we (i) quantify discrepancies between event time and information availability across MIMIC-IV workflows, (ii) identify temporal and semantic mismatches between commonly used features and prediction endpoints, (iii) distinguish patient-level from structural module-level missingness and examine their implications for imputation, and (iv) demonstrate these effects in mortality and phenotype-prediction case studies and derive a reproducible preprocessing checklist.

More broadly, we argue that preprocessing validity in EHR data is not an intrinsic property of a variable, timestamp, or modality, but depends on the modeling perspective, clinical episode, prediction anchor, and information state being represented. This yields three complementary forms of alignment considered throughout the paper: (i) temporal alignment, between event occurrence and information availability; (ii) semantic alignment, between the evidence contained in a data source and the claimed endpoint; and (iii) population alignment, between the population in which a feature or modality is observed and the population to which preprocessing is applied. MIMIC-IV serves as a case study for showing how violations of these alignments can remain hidden despite apparently reasonable predictive performance.

\section{Data}
\begin{figure}[h!]
    \centering
    \includegraphics[width=\linewidth]{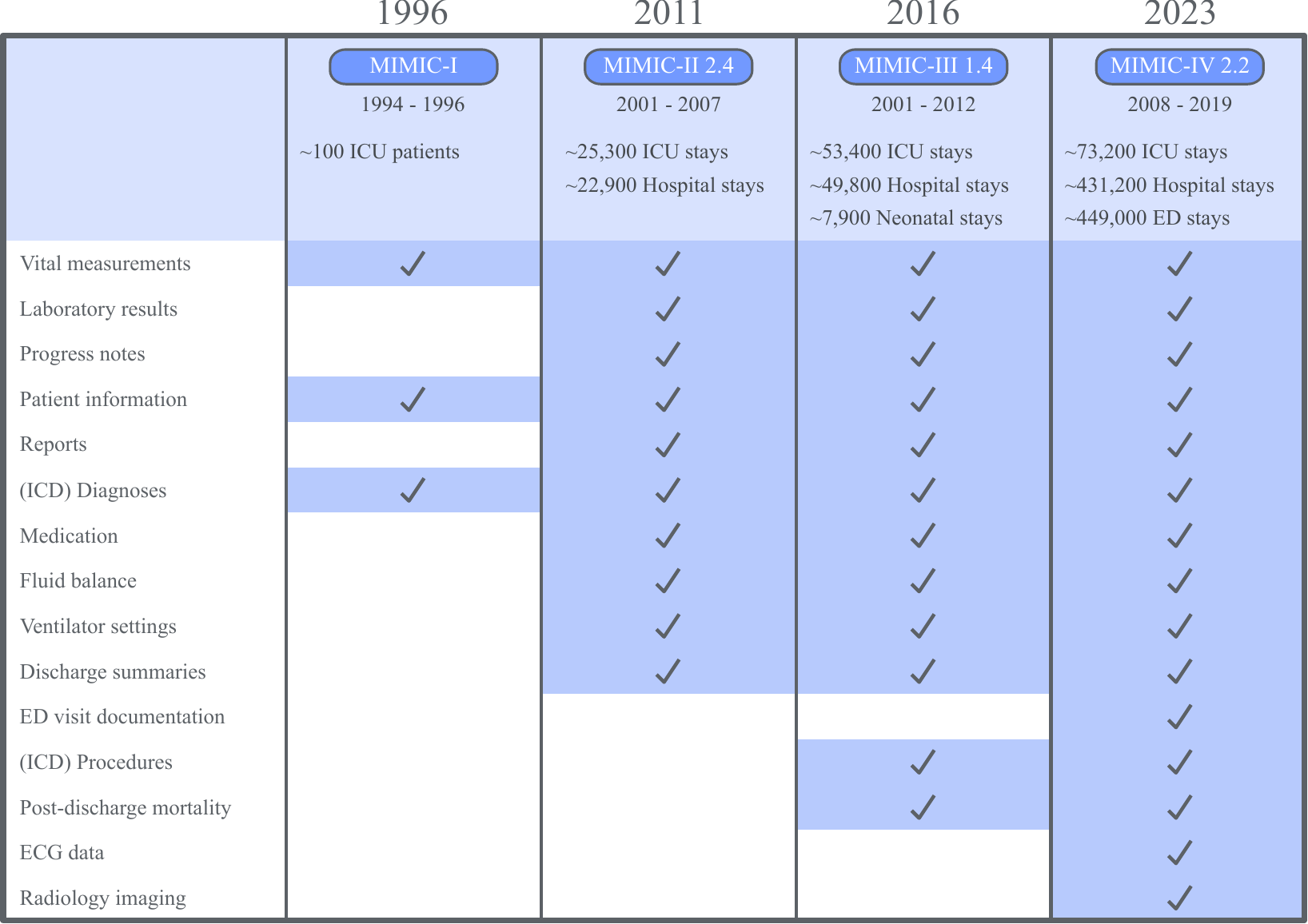}
    \caption{A broad overview of the different generations of the MIMIC dataset and the information categories they provide. The release dates, version numbers, and patient and visit counts correspond to the details documented in the official release publications accompanying each iteration. Information types shared across generations should not necessarily be interpreted as equivalent in scope, as the granularity, completeness, and source quality of the documented data have improved substantially across iterations. This paper's analyses use the latest version of MIMIC data available at the time of writing, i.e., MIMIC-IV 3.1. }
    \label{fig:mimic_versions}
\end{figure}

\begin{figure}[h!]
    \centering
    \includegraphics[width=\linewidth]{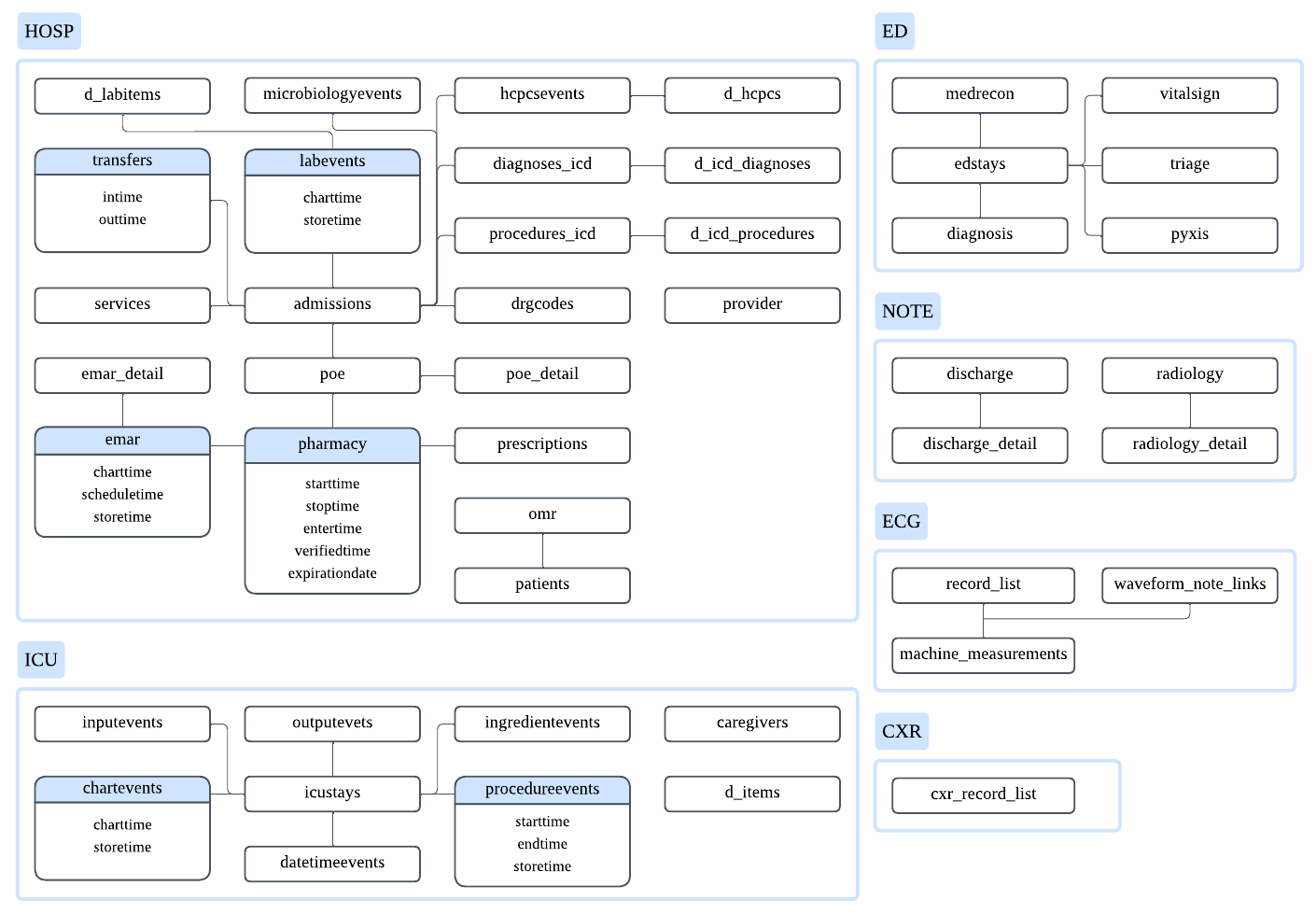}\caption{Overview of the table structure of the MIMIC-IV modules. The timestamps shown for the highlighted tables serve as representative examples of the different types of temporal information provided in MIMIC-IV. For the full overview of linkage identifiers and temporal fields relevant for episode alignment, refer to Figure \ref{fig:mimiciv_schema} in Appendix~\ref{app:mimic_schema}}
    \label{fig:mimic_table_condensed}
\end{figure}

The \textit{Medical Information Mart for Intensive Care} (MIMIC) is a publicly available, de-identified electronic health record database collected at Beth Israel Deaconess Medical Center. Four generations of the dataset have been released to date, with Fig. \ref{fig:mimic_versions} summarizing their scope and content. MIMIC-I ~\citep{Moody1996OG}, initially introduced in 1996, contains continuous physiological signals and clinical data recorded over a time period of 24 - 48 hours for approximately 100 ICU patients. Although the original design envisioned the inclusion of additional information from medical records, such as laboratory results, progress notes, and medications, these elements were later described as unavailable in MIMIC-I by its successor, MIMIC-II ~\citep{Saeed2011MIMIC-II}, which was published in 2011. MIMIC-II substantially expanded the dataset by introducing detailed digital therapeutic data of entire ICU stays and including laboratory measurements, medications, fluid balance, progress notes, and reports for more than 25,000 ICU stays between 2001 and 2007. MIMIC-III ~\citep{Johnson2016MIMICIII} further extended the dataset to more than 40,000 ICU patients spanning 2001 to 2012, adding procedure codes, post-discharge mortality, and improved structured documentation. It also included a few thousand neonatal patients, who were subsequently removed in MIMIC-IV. The latest generation, MIMIC-IV ~\citep{Johnson2023MIMIC}, covers admissions between 2008 and 2019 and expands beyond the ICU to include emergency department encounters. In contrast to previous releases, MIMIC-IV follows a modular structure that supports linkage to rich complementary datasets. The module MIMIC-CXR ~\citep{Johnson2019MIMICCXR, PhysioNet-mimic-cxr-jpg-2.1.0} contains chest X-rays and radiology reports; MIMIC-IV-ED ~\citep{PhysioNet-mimic-iv-ed-2.2} includes emergency department visit documentation; MIMIC-IV-ECG ~\citep{PhysioNet-mimic-iv-ecg-1.0} provides diagnostic electrocardiograms for a subgroup of patients; and MIMIC-IV-Note ~\citep{PhysioNet-mimic-iv-note-2.2} contains de-identified free-text clinical notes. Figure \ref{fig:mimic_table_condensed} provides an overview of all available MIMIC-IV modules, the tables within each module, and examples of the different temporal information available.

For a more detailed overview, Figure~\ref{fig:mimiciv_schema} in Appendix~\ref{app:mimic_schema} presents a visual schema including linkage identifiers and temporal fields relevant to episode alignment for all tables, while Figure~\ref{fig:mimiciv_schema_ext1} and Figure~\ref{fig:mimiciv_schema_ext2} show extended versions of the schema that also include table contents.

All analyses and reported results in this study are based on the following dataset versions: MIMIC-IV 3.1. ~\citep{PhysioNetMimiciv3.1}, MIMIC-IV-Note 2.2 ~\citep{PhysioNet-mimic-iv-note-2.2}, MIMIC-CXR 2.1.0 ~\citep{Johnson2019MIMICCXR, PhysioNet-mimic-cxr-jpg-2.1.0}, MIMIC-IV-ECG 1.0 ~\citep{PhysioNet-mimic-iv-ecg-1.0} and MIMIC-IV-ED 2.2 ~\citep{PhysioNet-mimic-iv-ed-2.2}. \\

The breadth and depth of MIMIC-IV have made it a dataset of substantial interest across diverse research areas of machine learning and clinical informatics. However, preprocessing a dataset of this scale and complexity can pose considerable challenges. Existing pipelines can reduce development time and facilitate comparable benchmark results, but they may also contain errors. Although our analysis primarily focuses on preprocessing pitfalls in MIMIC-IV pipelines, given that MIMIC-IV is the most recent and extensive generation of the MIMIC database, we also incorporate unaddressed preprocessing errors identified in pipelines for MIMIC-III, which continue to be widely used for benchmarking. By considering both generations, our objective is to provide a practical guide for researchers who either develop new preprocessing pipelines or select and validate existing benchmarks derived from the MIMIC dataset.

\section{Preprocessing Pitfalls} \label{sec:prep}

We organize the identified challenges according to the temporal, semantic, and population-alignment perspective introduced above, while using four operational failure modes to distinguish how misalignment enters a preprocessing pipeline. Throughout the paper, we use the following definitions:
\begin{itemize}[nosep]
    \item \textbf{Data leakage}: Information unavailable at the stated prediction time or outside of the defined clinical episode enters the model input, or information from the same patient or clinical episode crosses the train–test boundary.
    \item \textbf{Temporal target misalignment}: The endpoint refers to a different time interval than the predictors.
    \item \textbf{Semantic feature–target misalignment}: The selected feature or modality cannot plausibly provide evidence for the target at the intended prediction time, or the endpoints conflate distinct clinical phenomena. 
    \item \textbf{Selection/Imputation bias}: Preprocessing changes the represented population or inserts non-neutral values, for example when statistics estimated from an observed subgroup are transferred to systematically different unobserved cases.
\end{itemize}

\subsection{Temporal Alignment and Leakage} \label{temp}
 Preprocessing choices in MIMIC-IV can introduce several forms of temporal leakage into the data. First, a longitudinal form of \textbf{data leakage} may occur when random splits allocate multiple admissions from the same patient or multiple ICU stays within the same hospital admission across the training and test sets ~\citep{Kapoor2023Leakage, Sauer2022EHRPitfalls}. This can lead to overly optimistic performance estimates because information about the same patient may be seen during training. To mitigate this risk, data should be split by \texttt{subject\_id}. Alternatively, the 58.1\% of ICU stays with a previous admission, or the 9.8\% of repeated ICU admissions within the same stay, can be excluded from the analysis, depending on the intended study design.
 
A related but distinct issue regarding \textbf{data leakage} concerns the definition of the clinical episode of interest itself. Emergency department stays, hospital admissions, and ICU stays are distinct but temporally overlapping concepts, and their boundaries may not align due to different administrative documentation processes. For example, in almost all cases of hospital admissions linked to an ED stay, the hospital \texttt{admittime} precedes ED departure, with a median of 1.5 hours, meaning that the administrative admission boundary falls within the ED stay. Similarly, 19.7\% of ICU stays end after the documented hospital discharge time of the corresponding admission, with a median offset of 1 hour between the two timestamps. For ED stays that result in hospital admission, diagnostic evidence such as laboratory results or radiology imaging may be available and documented before the recorded hospital admission time. For instance, 21.6 \% of radiology reports have a \texttt{charttime} before hospital \texttt{admittime} but after ED \texttt{intime}, and may therefore constitute valid evidence depending on the chosen criteria. Defining a prediction episode, therefore, requires an explicit choice of anchor: which clinical abstraction is used, which observations are included, at what point within the episode the prediction is made, and how or whether features from other abstractions are temporally aligned with that anchor. 
 
\begin{figure}[t]
    \centering
    \includegraphics[width=\linewidth]{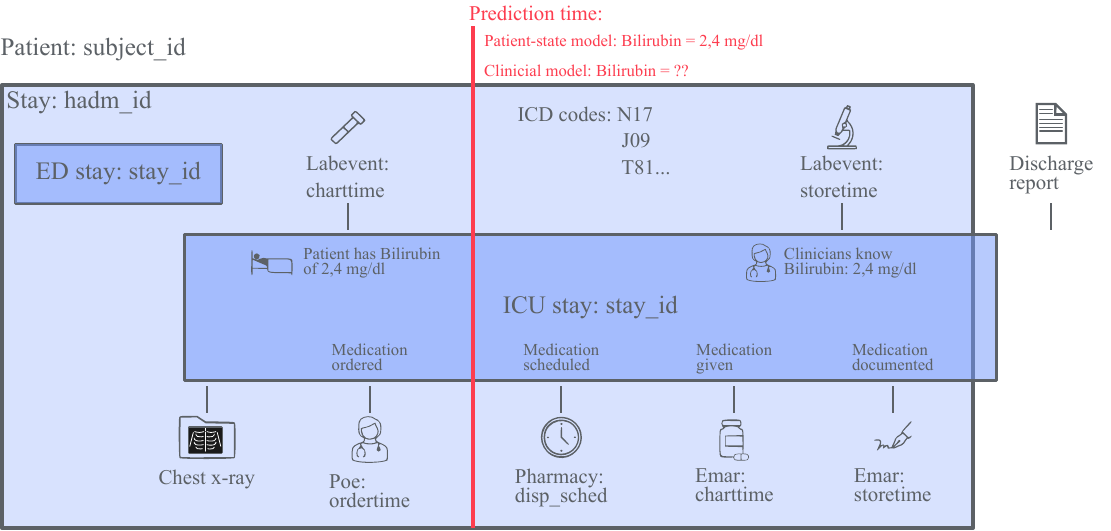}
    \caption{A conceptual illustration showing how information from different modules reconstructs a patient's clinical timeline. The overlap between ED and ICU stays, as well as the misaligned end of stay, shows discrepancies between documented admission and discharge times across stations. Because the discharge report is written retrospectively, its final timestamp can fall outside the documented hospital stay. ICD codes, for which no precise timestamp is documented, can only be matched to the respective stay. The figure highlights how different timestamps represent different temporal perspectives: for example, \texttt{charttime} can support patient-state modeling, whereas \texttt{storetime} provides a later proxy for information availability in clinical decision-making. Due to the temporal difference, the choice of timestamps can affect the information available at prediction time.}
    \label{fig:mimic_flow}
\end{figure}

 Temporal \textbf{data leakage} or \textbf{temporal target misalignment} can also occur within a single defined episode, particularly in prediction tasks based on the first 24 or 48 hours after ICU admission. These tasks are common in EHR-based analyses, with endpoints including length of stay ~\citep{Harutyunyan2019MIMICMultiTask, Zheng2025RealTimePred, Hempel2023LOS, Wang2020MIMICExtract} and in-hospital mortality ~\citep{Gupta2022AnEDA, Harutyunyan2019MIMICMultiTask, Meng2022DLInterpret, Zheng2025RealTimePred, Wang2020MIMICExtract}. A common pitfall for \textbf{data leakage} is the inclusion of information that would only have become available after the defined observation window during preprocessing. Some studies have acknowledged this limitation explicitly, for example by noting the inclusion of the last chest X-ray of the entire stay in otherwise 48-hour-based input data ~\citep{hayat2023medfuse}. A more frequent issue, however, is the use of ICD codes (International Classification of Diseases), for which no event-level timestamps are available. Within a defined episode, leakage commonly arises from stay-level ICD codes without event timestamps, ambiguous timestamp semantics in tabular data, and retrospective discharge notes.
 Sections 3.1.1–3.1.3 examine these three cases in turn. Figure~\ref{fig:mimic_flow} complements the following analyses by providing a conceptual view of a patient's timeline during a stay, illustrating the temporal level and order in which information can be reconstructed from the data of multiple modules.

\subsubsection{Temporal Uncertainty of ICD Codes} \label{sec:temporal_uncertainty_ICD}
 ICD codes encode diagnoses and procedures, and are commonly used to define clinical subgroups ~\citep{Gupta2022AnEDA}, serve as model features ~\citep{Hempel2023LOS}, or as labels for prediction tasks, including common phenotypes ~\citep{Gupta2022AnEDA, Harutyunyan2019MIMICMultiTask, woznica2022consolidated} and specific disease groups ~\citep{Liang2021multibench, Dufumier2024CoMM, Liang2023FactorCL}. However, an assumption encountered in some MIMIC-based analyses is that encoded diagnoses represent conditions already present at admission, including the claim that ``MIMIC only records diagnosis at the start of every hospital admission'' ~\citep{Gupta2022AnEDA}. In reality, ICD codes retrospectively summarize conditions and procedures over the entire hospital stay of a patient for billing purposes, including complications and hospital-acquired infections that may emerge over the course of treatment ~\citep{Sauer2022EHRPitfalls}. MIMIC, however, does not provide timestamps for individual ICD code assignments, meaning they can only be attributed to the stay as a whole. Table \ref{tbl-icd-leakage} presents a selection of codes that correspond to diagnoses such as post-procedural complications or known antibiotic resistance, which are unlikely to have been clinically evident at admission time. These examples support the interpretation that ICD codes in MIMIC-IV reflect information accumulated over the entire stay. For most codes in the dataset, however, there are no clear lexical indicators of whether the corresponding condition was present at the admission time, particularly when codes are grouped into coarse-grained categories based solely on their initial digits. The assumption that ICD codes represent admission-time knowledge can therefore introduce \textbf{data leakage} or \textbf{temporal target misalignment}.

In some cases, this assumption may have a limited impact. For example, \cite{Gupta2022AnEDA} use the final hours of a patient's stay to predict phenotype labels for the next admission; in such a setting, filtering patients according to certain chronic diseases during the current stay does not necessarily induce temporal information leakage. In other scenarios, however, the risk of temporal \textbf{data leakage} or \textbf{temporal target misalignment} is substantially higher. When ICD codes are used as model features ~\citep{Hempel2023LOS}, stay-level codes may expose the model to information that would not have been available at the prediction time, leading to \textbf{data leakage}. When ICD code endpoints are predicted from the first 24 or 48 hours of data ~\citep{Liang2021multibench, Dufumier2024CoMM, Liang2023FactorCL}, the absence of code-level timestamps creates \textbf{temporal misalignment} between target and features that injects noise into the training objective, as the exact time of disease onset, its duration, and whether the patient showed any clinical signs of the predicted disease during the observation window remain unknown. As a result, the model may be unintentionally incentivized to infer the patient's overall condition or risk of future complications rather than to identify clinical evidence of an already existing disease. This not only compromises the model's performance but also undermines its potential for providing pathophysiologically meaningful associations and interpretable predictions, thereby limiting the clinical interpretability and validity of the resulting predictions unless the endpoint is explicitly formulated as a future or stay-level outcome.

\begin{table*}
\caption{Examples of ICD codes that are likely to occur during a hospital stay and their frequency in the MIMIC-IV 3.1 ~\citep{PhysioNetMimiciv3.1} dataset.}\label{tbl-icd-leakage}
\resizebox{\textwidth}{!}{
\begin{tabular}{llr}
\toprule
 ICD Code & Description of Code & Count\\ 
\midrule
\rowcolor{gray!8}
 998.1&Other complications of procedures not elsewhere classified: & 4808 \\
 \rowcolor{gray!8}
 &Hemorrhage or hematoma complicating a procedure not elsewhere classified& \\
 
 T82.8 &Complications of cardiac and vascular prosthetic devices, implants and grafts: & 4712 \\
 &Other specified complications of cardiac and vascular prosthetic devices, implants and grafts&\\

 \rowcolor{gray!8}
 996.7 &Complications peculiar to certain specified procedures:& 4494 \\
 \rowcolor{gray!8}
 &Other complications due to internal prosthetic device, implant, and graft&\\
 
 998.5 &Other complications of procedures, not elsewhere classified: Postoperative infection, not elsewhere classified & 3995 \\

 \rowcolor{gray!8}
 996.8&Complications peculiar to certain specified procedures: Complications of transplanted organ &3570 \\
 
 T81.4&Complications of procedures, not elsewhere classified: Infection following a procedure& 3293\\

 \rowcolor{gray!8}
 996.6&Complications peculiar to certain specified procedures:&3276\\
 \rowcolor{gray!8}
 &Infection and inflammatory reaction due to internal prosthetic device implant and graft&\\

 J95.8& Intraoperative and postprocedural complications and disorders of respiratory system, not elsewhere classified:   & 2970\\
 &Other intraoperative and postprocedural complications and disorders of respiratory system,& \\
 & not elsewhere classified& \\

 \rowcolor{gray!8}
 997.4&Complications affecting specified body system, not elsewhere classified: &2814\\
 \rowcolor{gray!8}
 &Digestive system complications, not elsewhere classified& \\
 
 K91.8 &Intraoperative and postprocedural complications and disorders of digestive system, not elsewhere classified:& 2549 \\
 &Other intraoperative and postprocedural complications and disorders of digestive system& \\

\rowcolor{gray!8}
 997.1 & Complications affecting specified body system, not elsewhere classified: & 2134 \\
 \rowcolor{gray!8}
 &Cardiac complications, not elsewhere classified &\\
 
 Z16.2 &Resistance to antimicrobial drugs: Resistance to other antibiotics & 1967\\

\rowcolor{gray!8}
 997.3 & Complications affecting specified body system, not elsewhere classified: & 1535\\
 \rowcolor{gray!8}
 &Respiratory complications, not elsewhere classified&\\
 
 T81.3 &Complications of procedures, not elsewhere classified: Disruption of wound, not elsewhere classified& 1449\\

 \rowcolor{gray!8}
 T83.5 &Complications of genitourinary prosthetic devices, implants and grafts:& 1315 \\
 \rowcolor{gray!8}
 &Infection and inflammatory reaction due to prosthetic device, implant and graft in urinary system&\\

 999.3 &Complications of medical care, not elsewhere classified: & 1311 \\
 & Other infection due to medical care, not elsewhere classified&\\

 \rowcolor{gray!8}
 I97.8 &Intraoperative and postprocedural complications and disorders of circulatory system, not elsewhere classified:& 1293 \\
 \rowcolor{gray!8}
 &Other intraoperative and postprocedural complications and disorders of the circulatory system, & \\
 \rowcolor{gray!8}
 &not elsewhere classified&\\

 998.3 &Other complications of procedures, not elsewhere classified: Disruption of operation wound & 1177\\

\rowcolor{gray!8}
 L76.3 &Intraoperative and postprocedural complications of skin and subcutaneous tissue: & 1143 \\
 \rowcolor{gray!8}
 &Postprocedural hematoma and seroma of skin and subcutaneous tissue following a procedure&\\

 998.2 &Other complications of procedures, not elsewhere classified: & 1103 \\
 &Accidental puncture or laceration during a procedure, not elsewhere classified&\\

\rowcolor{gray!8}
 998.8 &Other complications of procedures, not elsewhere classified:& 994 \\
 \rowcolor{gray!8}
 &Other specified complications of procedures, not elsewhere classified& \\

 I97.1 &Intraoperative and postprocedural complications and disorders of circulatory system, not elsewhere classified:& 888 \\
 &Other postprocedural cardiac functional disturbances&\\

 \rowcolor{gray!8}
 T81.1& Complications of procedures, not elsewhere classified: Postprocedural shock&814\\
\bottomrule
\end{tabular}
}
\end{table*}

To avoid \textbf{temporal target misalignment} or \textbf{data leakage} when using ICD codes, the prediction task can be defined at the end of the entire hospital stay ~\citep{Harutyunyan2019MIMICMultiTask}. Alternatively, studies should restrict inputs to features with clearly documented timestamps, avoid using stay-level ICD codes as time-specific features, and apply careful feature selection based on whether each variable could plausibly have been known at the intended prediction time.

\subsubsection{Temporal Ambiguity in Tabular Data}
\label{sec:temporal_ambiguity_tabular_data}


\begin{figure}[t]
    \centering
    \includegraphics[width=\textwidth]{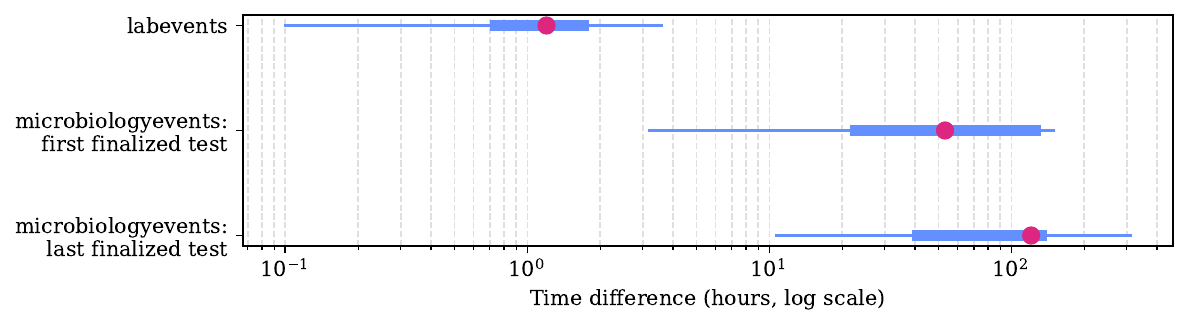}
    \caption{Analysis of the time difference in hours on a log scale between \texttt{charttime} and \texttt{storetime} in the \texttt{labevents} and \texttt{microbiologyevents} tables. Median is denoted by a dot, 25th to 75th percentile by a thick line, and 5th to 95th percentile by a thin line. Exact median and 5th to 95th percentile are 1.2 hours (0.1--3.6) for labevents, 53.2 hours (3.2--149.1) for the first finalized test of microbiologyevents, and 120.7 hours (10.7--311.5) for the last finalized test of microbiologyevents.}
    \label{fig:chart-store-leakage}
\end{figure}

For tabular data, timestamps are available and can therefore support temporal alignment. However, the mere presence of timestamps does not by itself prevent \textbf{data leakage}. Many MIMIC-IV tables contain multiple timestamps for a single clinical process, and these timestamps often correspond to different stages of the workflow, as visualized in Figure \ref{fig:mimic_flow}. A preprocessing pipeline can therefore be temporally invalid even when all selected rows appear to occur before the chosen prediction horizon.

A central distinction must be made between \textit{event occurrence} and \textit{information availability}. For events in \texttt{chartevents} and \texttt{datetimeevents}, \texttt{charttime} reflects the time at which an observation was made or measured, whereas \texttt{storetime} reflects when the event was documented or validated by clinical staff. For free-text notes, the associated tabular timestamp \texttt{charttime} refers to the time at which the note was charted, while \texttt{storetime} refers to the time at which the completed and signed note was stored. For \texttt{labevents} and \texttt{microbiologyevents}, the distinction is even more consequential: \texttt{charttime} typically anchors specimen acquisition or measurement time, whereas \texttt{storetime} records when the result was entered into the system and therefore provides a later proxy for potential information availability. An overview of all timestamps of each table is given in Figure \ref{fig:mimiciv_schema} in Appendix \ref{app:mimic_schema}.
Our analysis of documented items in Figure \ref{fig:chart-store-leakage} shows that in \texttt{labevents}, \texttt{storetime} is documented with a median delay time of 1.2 hours relative to \texttt{charttime}. This produces a meaningful temporal alignment shift in the modeled stay trajectory, even when 1-hour bins are used. In \texttt{microbiologyevents}, this temporal shift is substantially larger. For a given specimen, multiple tests can be ordered, and the median delay between \texttt{charttime} and \texttt{storetime} of the first finalized test is 53.2 hours. The delay until all specimen tests are finalized is even longer, with a median of 120.7 hours. Due to this delay, 24.7\% of final microbiology test results become available only after patient discharge. Thus, using \texttt{charttime} alone can be appropriate when the modeling goal is to reconstruct the patient's biological state at the time of measurement. It is generally insufficient when the model is intended to approximate what information could have been available to a clinician or hospital information system at a specific decision time. 

This distinction is especially clear for laboratory data. A blood test may describe the patient's physiological state at the time the specimen was drawn, but the corresponding result is not known until laboratory processing has been completed and the result has been stored ~\citep{Sauer2022EHRPitfalls}. 
For decision-support models, laboratory features should therefore generally satisfy both \texttt{charttime} $\leq t_{\mathrm{pred}}$ and \texttt{storetime} $\leq t_{\mathrm{pred}}$ when \texttt{storetime} is available as a proxy for result availability; patient-state models may instead anchor observations to specimen time. The same distinction applies to microbiology, where specimen collection, preliminary results, organism identification, susceptibility testing, and finalization represent different information states.

Medication data introduces an additional layer of ambiguity because timestamps refer not only to time but also to workflow state. The tables \texttt{poe}, \texttt{pharmacy}, \texttt{prescriptions}, and \texttt{emar} represent different stages of the medication process. \texttt{poe} records provider order entry and therefore primarily represents treatment intent. \texttt{pharmacy} and \texttt{prescriptions} contain prescription and pharmacy workflow information, including \texttt{starttime}, \texttt{stoptime}, \texttt{entertime}, \texttt{verifiedtime} and \texttt{disp\_sched}. These timestamps may describe planned treatment intervals or pharmacy verification rather than actual medication administration. \texttt{emar} is closer to bedside administration documentation, with \texttt{charttime} indicating the documented administration or medication-status time and \texttt{storetime} indicating documentation time. However, eMAR data are not available for all patients, as the eMAR system was introduced during part of the data collection period. As a result, 44.9\% of admissions with pharmacy information have no eMAR documentation. Among \texttt{pharmacy} entries without a match in \texttt{emar}, 36.9\% belong to admissions where eMAR documentation exists, but no \texttt{emar} entry can be matched to the corresponding \texttt{pharmacy\_id}. This may be partly explained by changes in the patient's condition that caused a planned treatment to be discontinued, or by linkage mismatches between the two tables. Moreover, \texttt{emar} may document that a dosage was not given or may provide no specific administration-status information in the \texttt{event\_txt} field. Accordingly, only 69.5\% of \texttt{emar} entries confirm likely medication administration, whereas 15.5\% explicitly deny administration. 
These issues reduce the fraction of admissions and medications for which actual exposure can be confirmed using the more precise eMAR documentation, but for most downstream clinical prediction tasks that require actual treatment exposure, \texttt{emar} remains preferable as a closer approximation to bedside administration. Given these differences in granularity and workflow semantics, medication features should always explicitly state whether they represent order intent, planned prescription, verified pharmacy workflow, documented administration, or inferred exposure from scheduled administration times.

The appropriate timestamp to prevent \textbf{data leakage}, therefore, depends on the modeling goal. We distinguish two broad categories. The first is \textit{patient-state or exposure modeling}. Here, the goal is to describe what happened to the patient during a clinically meaningful window: for example, the patient's physiological state, pathogen state, biomarker state, or treatment exposure. In this setting, event timestamps such as \texttt{charttime}, specimen collection time, or \texttt{starttime} are defensible, provided the study does not claim that the information was available to clinicians at that time. The second category is \textit{clinician-knowledge or decision-support modeling}. Here, the goal is to emulate or support clinical decisions at a specific prediction anchor. In this setting, features should be restricted to information that could reasonably have been available by the prediction time according to the available workflow and timestamp proxies. When the correct timestamp is ambiguous, the conservative choice is to use the later timestamp or to exclude the feature.

Several prior studies are best interpreted as patient-state or exposure analyses and are therefore not automatically invalid even when they rely on event-time semantics. In these cases, the relevant question is whether a physiological state, biomarker state, pathogen state, or treatment exposure occurred within a clinically meaningful window, not necessarily whether the information was available to a clinician at the same time. For studies of organism type in sepsis ~\citep{guo2023organism}, the C-reactive-protein--triglyceride--glucose index~\citep{wang2025cti}, the triglyceride-glucose index in critically ill cardiovascular-kidney-metabolic patients~\citep{tygckm2025}, the lactate-to-calcium ratio in sepsis-induced myocardial injury~\citep{dou2026lcr}, and serum bicarbonate in critically ill influenza patients~\citep{serumbicarb2026}, event-time semantics are compatible with a retrospective patient-state or first-day biomarker interpretation; if interpreted instead as early clinician-available prediction, result-availability filtering would additionally be required. 

Similarly, first-day ICU mortality models~\citep{sun2023cardiacarrest,earlyicu2024} can be interpreted as 24-hour landmark models; an admission-time interpretation would additionally require laboratory availability and treatment-workflow semantics to be enforced. For treatment-exposure studies, such as corticosteroid analyses in sepsis~\citep{corticosteroids2023sepsis}, the temporal interpretation depends on whether exposure is reconstructed from actual administration rather than from orders, prescriptions, or pharmacy intervals. In contrast, the use of discharge notes for post-discharge prediction~\citep{lcd2024} is temporally aligned when the prediction anchor is discharge or note completion, whereas the same notes would not support an admission-time prediction interpretation.

Availability requirements are stricter for clinician-knowledge, triage, early-prediction, and decision-support models. Under its stated emergency-department decision-support framing during the first 1.5 hours after ED arrival, the publicly described preprocessing of MDS-ED / MIMIC-IV-Ext-MDS-ED~\citep{lopezalcaraz2025mdsed} appears to permit laboratory information whose availability before the prediction anchor cannot be established, because events are filtered by \texttt{labevents.charttime} without also enforcing \texttt{storetime}. For the MIMIC-IV-ED benchmark~\cite{xie2022mimicived}, temporal validity depends on the prediction task: later ED-stay features, such as last vital signs or medication counts, would not be appropriate for a triage-time interpretation but are compatible with post-ED or ED-disposition tasks in which full ED-stay information is intended. Under a strict early clinician-knowledge interpretation, event-time filtering in the multimodal risk-prediction pipeline~\citep{wang2024multimodalrisk} does not by itself establish that laboratory values, microbiology results, radiology reports, and notes were available within the first 48 hours.

Several other decision-support resources remain temporally ambiguous rather than definitively leaking. For MIMIC-IV-Ext-CDM~\citep{hager2024cdm}, ARES / ETHOS~\citep{renc2025ethos}, the first-24-hour ICU mortality model by \cite{chen2025hypertensionaf}, hourly ICU deterioration prediction with structured data and notes~\citep{sadanandan2026hourly}, and the initial-24-hour multimodal ICU mortality model~\citep{multimodalicu2025}, the accessible descriptions do not provide sufficient detail to verify availability-aware timestamp handling for all feature families. MIMIC-IV-Ext-CLIF~\citep{mimicivextclif} explicitly separates laboratory collection time from result availability, thereby supporting availability-aware downstream use, although temporally inappropriate use remains possible.

Some studies combine first-window structured features with later textual evidence, making the source of leakage mixed. For example, \cite{mamatov2025cardiacnotes} extract first-24-hour structured ICU variables, including vital signs, laboratory values, and treatment indicators. These first-24-hour structured variables are consistent with a 24-hour landmark interpretation; an earlier decision-time interpretation would require additional availability constraints. A separate temporal-availability concern arises from the use of discharge summaries for a task framed as early mortality prediction. This issue is discussed in the following section on discharge-note leakage.

Overall, studies should report the prediction anchor, source tables, timestamp columns, availability rules, and workflow interpretation for every feature family. When the appropriate timestamp cannot be determined, the safest preprocessing strategy is to use the later timestamp or to exclude the feature from early prediction.

\subsubsection{Temporal Evidence within Discharge Notes}
\label{sec:temporal_evidence_discharge_notes}
Discharge summaries are retrospective documents that summarize a patient's history, diagnostic workup, treatment course, and discharge status. In MIMIC-IV, they follow a recurring section structure, including sections such as \textit{Chief Complaint}, \textit{History of Present Illness}, \textit{Past Medical History}, \textit{Brief Hospital Course}, \textit{Physical Exam}, and \textit{Discharge Diagnosis} \citep{PhysioNet-mimic-iv-note-2.2}. This structure enables section-level preprocessing, but the temporal meaning of a section is not determined by its heading alone.

Table~\ref{tab:temporal_sections} summarizes common discharge-note sections according to four evidence windows: historical evidence, admission-time evidence, in-stay evidence, and discharge-time evidence. \textit{Historical evidence} includes sections such as \texttt{Sex}, \texttt{Allergies}, \texttt{History of Present Illness}, \texttt{Past Medical History}, \texttt{Social History}, and \texttt{Family History}. \textit{Admission-time evidence} includes \texttt{Service}, \texttt{Chief Complaint}, \texttt{Medications on Admission}, admission-related parts of \texttt{Physical Exam}, and admission-related information in the \texttt{Brief Hospital Course}. \textit{In-stay evidence} includes \texttt{Major Surgical or Invasive Procedure}, \texttt{Pertinent Results}, stay-related findings in \texttt{Physical Exam}, and course-related information in the \texttt{Brief Hospital Course}. \textit{Discharge-time evidence} includes discharge-related findings in \texttt{Physical Exam}, the final narrative synthesis in the \texttt{Brief Hospital Course}, \texttt{Discharge Medications}, \texttt{Discharge Disposition}, \texttt{Discharge Diagnosis}, and \texttt{Discharge Condition}. These categories are not mutually exclusive: \texttt{Physical Exam} and \texttt{Brief Hospital Course} can contain information from several phases of the admission.

Our analysis shows that the main section order is mostly stable, but not sufficient for automatic temporal filtering. Most major sections follow the expected order, whereas optional or less stable sections require special handling. For example, \texttt{Brief Hospital Course} appeared in only 88.0\% of sampled notes. We also observed repeated headings, empty headings followed by repeated non-empty headings, de-identified heading variants such as \texttt{\_\_\_ Complaint}, and heading-like strings inside section content. These cases can cause unrestricted regular-expression searches to select the incorrect section boundary.

Section-level comparisons with structured tables further show that overlap does not imply redundancy. The extracted discharge-note \texttt{Chief Complaint} matched or partially contained the ED triage chief complaint only in a subset of linked cases, even after abbreviation and synonym normalization. Thus, \texttt{ed.triage.chiefcomplaint} is useful for plausibility checking, but not as a gold standard. \texttt{Medications on Admission} overlaps with medication reconciliation data, but the note often contains richer details, such as dose, route, or narrative context. \texttt{Pertinent Results} overlaps strongly with other sources: laboratory findings often correspond to abnormal entries in \texttt{hosp.labevents}, microbiology findings overlap with \texttt{hosp.microbiologyevents}, and imaging subsections frequently summarize \texttt{Impression} statements from radiology reports. \texttt{Discharge Diagnosis} overlaps with ICD codes, but represents clinician-facing diagnostic language rather than standardized billing codes.

Prior work uses discharge-note sections with varying levels of methodological transparency. \cite{VanAken2021AdmissionNotes} and \cite{Rohr2024COPIV} use section-level filtering to construct admission-time prediction settings, making the temporal assumption explicit. \cite{Shing2021ClinicalEncounterSummarization} extract multiple MIMIC-III discharge-summary sections using section-header variants, extract content until the next section header, and manually inspect extracted sections. The CLIP dataset combines regular-expression labels for \texttt{Followup Instructions} and \texttt{Discharge Instructions} with physician annotation and revision ~\citep{Mullenbach2021CLIP}. \cite{He2024ShimoLab} describe regex-based extraction for simple and complex MIMIC-IV discharge-summary sections and report failures caused by inconsistent formatting. In contrast, several summarization and coding studies rely on regex-extracted sections as task-specific preprocessing, but report fewer details about aliases, repeated headings, optional sections, boundary validation, or failure cases \citep{Aali2024MIMICIVBHC,Xu2024DischargeMe,Lyu2024UFHOBI,Damm2024WisPerMed,Hartman2022HospitalCourse,Searle2023BHC,Wang2024DRGLLAMA,Bi2024HospitalCourse,Chung2025VeriFactBHC,Schafer2020MultilingualICD,Shi2017AutomatedICD,Teng2020GCoder}. Dedicated section-identification work supports this concern: clinical section identification is non-trivial, and regex segmentation can diverge from clinician-annotated section boundaries ~\citep{PomaresQuimbaya2019Sections,Landes2022MedSecId}.

A separate \textbf{temporal target misalignment} risk arises when discharge summaries are used without aligning the document type to the prediction anchor. \cite{mamatov2025cardiacnotes} illustrate this issue: the study is framed as early in-hospital mortality prediction for ICU cardiac-arrest patients while incorporating discharge summaries and radiology reports. Because discharge summaries are retrospective documents, their use is temporally aligned when the prediction anchor is discharge or note completion; for earlier prediction anchors, their availability cannot be assumed. Radiology reports can be temporally aligned with earlier prediction anchors when the report text was available before the prediction time.

For leakage-sensitive tasks, section splitting should satisfy six criteria: (i) use an explicit heading inventory with aliases, spelling variants, and de-identified variants; (ii) apply order-aware extraction using the next valid section boundary rather than unrestricted global search; (iii) handle optional, skipped, repeated, and empty sections explicitly; (iv) retain both extracted and residual text for auditing; (v) assign each extracted section a prediction-time validity status; and (vi) report failure cases, including missing sections, malformed headings, repeated headings, and sections whose content extends beyond the expected temporal window.

These criteria are also motivated by broader documentation studies. Electronic health records contain substantial redundancy and copied text, and discharge-summary provenance analyses show that clinicians often copy, summarize, or adapt information from prior documentation ~\citep{Wrenn2010Redundancy,Tsou2017CopyPaste,Landes2023Provenance}. The main risk is therefore not only missing a heading, but also extracting content that is formally located under one section while semantically or temporally belonging to another part of the patient trajectory, adding \textbf{data leakage}.

\begin{table*}[tb]
\centering
\small
\renewcommand{\arraystretch}{1.12}
\setlength{\tabcolsep}{6pt}
\caption{Temporal organization of sections in MIMIC-IV discharge notes. An \textbf{x} indicates that the section typically contains information connected to the corresponding temporal interval.}
\label{tab:temporal_sections}

\begin{tabular}{
p{0.34\linewidth}
>{\centering\arraybackslash}p{0.13\linewidth}
>{\centering\arraybackslash}p{0.13\linewidth}
>{\centering\arraybackslash}p{0.13\linewidth}
>{\centering\arraybackslash}p{0.13\linewidth}
}
\toprule
Section
& \multicolumn{4}{c}{Temporal Connection} \\
\cmidrule(lr){2-5}
& Historical
& At Admission
& During the Stay
& At Discharge \\
\midrule

\rowcolor{gray!08}
Sex                                      & $\times$  &   &   &   \\
Service                                  &   & $\times$ &   &   \\
\rowcolor{gray!08}
Allergies                                & $\times$  &   &   &   \\
Chief Complaint                          &   & $\times$ &   &   \\
\rowcolor{gray!08}
Major Surgical or Invasive Procedure     &   &   & $\times$ &   \\
History of Present Illness               & $\times$ &   &   &   \\
\rowcolor{gray!08}
Past Medical History                     & $\times$ &   &   &   \\
Social History                           & $\times$ &   &   &   \\
\rowcolor{gray!08}
Family History                           & $\times$ &   &   &   \\
Physical Exam                            &   & $\times$ & $\times$ & $\times$ \\
\rowcolor{gray!08}
Pertinent Results                        &   &   & $\times$ &   \\
Brief Hospital Course                    & $\times$  & $\times$ & $\times$ & $\times$ \\
\rowcolor{gray!08}
Medications on Admission                 &   & $\times$ &   &   \\
Discharge Medications                    &   &   &   & $\times$ \\
\rowcolor{gray!08}
Discharge Disposition                    &   &   &   & $\times$ \\
Discharge Diagnosis                      &   &   &   & $\times$ \\
\rowcolor{gray!08}
Discharge Condition                      &   &   &   & $\times$ \\

\bottomrule
\end{tabular}
\end{table*}

\subsection{Semantic Alignment and Feature-Task Misalignment}
\begin{figure*}[t]
    \centering
    \includegraphics[width=\textwidth]{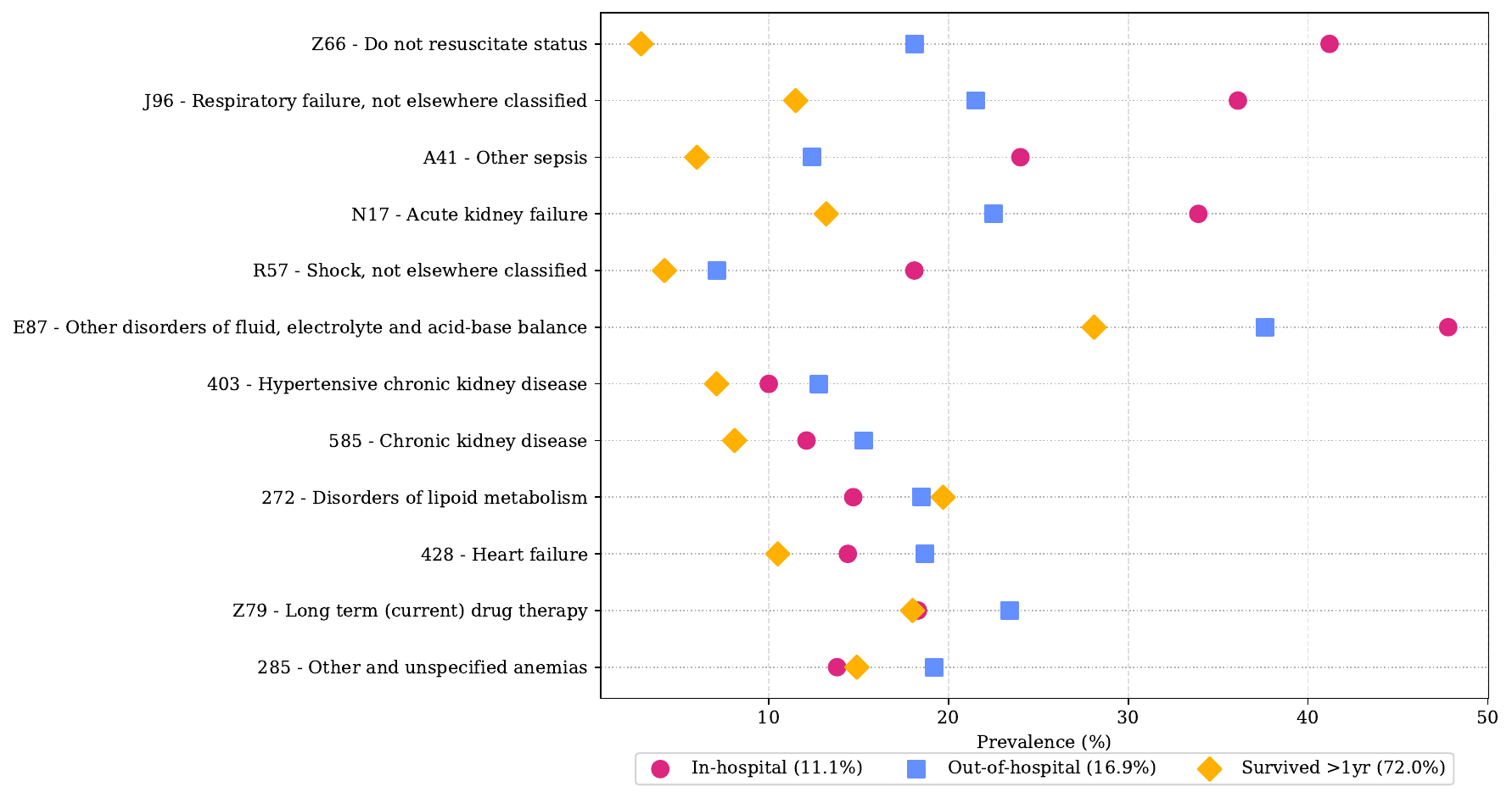}
    \caption{Differences in ICD code prevalence between mortality subgroups of ICU stays. Percentages for ICD code occurrences are reported relative to the full subgroup; in-hospital includes 9,475 stays (11.1\%), out-of-hospital 14,383 stays (16.9\%), and survived $>1$yr 61,384 stays (72.0\%).}
    \label{fig:ICU-mort}
\end{figure*}

\textbf{Semantic feature–target misalignment} arises when endpoints conflate distinct clinical phenomena or when selected features and modalities provide weak or inconsistent evidence for the target. 
A representative example is mortality prediction, where a distinction should be made between in-hospital and out-of-hospital mortality. In-hospital mortality is commonly predicted using data collected during the first 24–48 hours of ICU admission ~\citep{Gupta2022AnEDA, Harutyunyan2019MIMICMultiTask, Meng2022DLInterpret, Zheng2025RealTimePred}. In contrast, out-of-hospital mortality can benefit from information accumulated over the entire hospital stay ~\citep{Gao2025Mortality} and reflects a clinically distinct endpoint. Out-of-hospital mortality is also noisier because it is often assessed over a one-year follow-up period and may include deaths unrelated to the index hospitalization. From a medical perspective, these endpoints may follow different causal pathways ~\citep{Gao2025Mortality}, making their separation advisable. In MIMIC-IV, this distinction is reflected in the prevalence of different ICD codes across the groups. As shown in Figure \ref{fig:ICU-mort}, acute critical and explicit wishes against resuscitation are more frequent among patients with in-hospital mortality, whereas chronic diseases of the kidney, heart failure, and long-term medical treatments are more frequent among patients with out-of-hospital mortality.

While combining these endpoints and predicting overall mortality from early admission data  ~\citep{Liang2021multibench} is not strictly invalid, such models may be less directly clinically actionable because they combine endpoint-specific mechanisms and make it more difficult to determine which clinical process drives the prediction.

A related problem arises when a nominally single modality is inherently heterogeneous. The MIMIC-IV-Note module is already separated into discharge letters and radiology reports according to the native table configuration. However, the radiology reports themselves cover diverse imaging procedures and body regions. For use cases that require distinguishing between imaging types ~\citep{LAM2025PERadiology, Hager2024LLMClinDec}, MIMIC-IV-Note does not provide explicit modality labels, necessitating the extraction of information from the text. In such cases, regular expressions should first be used to identify the descriptor line indicating which imaging study was performed, and only this line should be used to infer the procedure. A naive full-text analysis can mistakenly treat references to other imaging procedures as the performed modality, resulting in a 52.6\% difference in modality assessment compared to a descriptor-based analysis.

MIMIC is also widely used for multimodal learning, but the value of additional modalities depends on their temporal and semantic relevance to the prediction endpoint.
For example, chest X-rays from the MIMIC-CXR module ~\citep{Johnson2019MIMICCXR} are used in some studies as a second modality, but do not always match the prediction tasks well. One endpoint observed in the literature ~\citep{hayat2023medfuse, Rasekh2024Fusion, Yao2024DrFuse} is the classification of 25 common phenotypes based on the ICD codes assigned during the hospital stay. Yet only a small subset of these phenotypes are primarily diagnosed using chest X-rays, including pneumonia, chronic obstructive pulmonary disease, and pleurisy. For some phenotypes, such as cerebrovascular disease, respiratory failure, or chronic kidney disease, chest X-rays may provide indirect or incidental evidence depending on the clinical presentation and severity at the time of imaging. For others, such as liver diseases, diabetes mellitus, or septicemia, the diagnosis relies entirely on other clinical, laboratory, or imaging procedures. Consequently, chest X-rays can provide only limited relevant information for many of the 25 classification targets, leading to \textbf{semantic feature-target misalignment}. 

We encountered a similar issue when implementing a binary classification task based on whether a code from the respiratory disease category was assigned during hospitalization. Although respiratory diseases and chest X-rays may appear to be inherently linked, this ICD category also includes all diseases of the upper respiratory tract. As most of the upper respiratory tract is not physically captured in chest X-ray imaging, these conditions may leave no direct visual evidence in the image. This creates \textbf{semantic feature-target misalignment}: the label indicates the presence of a respiratory disease, while the chest X-ray may appear normal with respect to that condition, thereby introducing noise into the training process. Additionally, as discussed in Section \ref{temp}, the disease used as the endpoint may not have been present at the time the features were collected, introducing additional \textbf{temporal target misalignment} and further increasing noise. 

In contrast, \cite{Soenksen2022mimiciv} demonstrated performance improvements from including chest X-rays compared to unimodal models across all multimodal tasks in their HAIM framework. Their HAIM-MIMIC-MM dataset combines four modalities: demographic tabular data, time series, chest X-rays, and notes. Two of their twelve predictive tasks concern the overall patient state, namely 48-hour mortality and length of stay, while ten target the presence or absence of chest pathologies. Their careful task design avoids much of the modality-endpoint mismatch and temporal uncertainty described above by selecting conditions for disease prediction tasks that can be diagnosed with chest X-rays in the MIMIC-CXR-JPG dataset ~\citep{PhysioNet-mimic-cxr-jpg-2.1.0}. The labels provided by MIMIC-CXR-JPG are derived directly from the radiology notes associated with each image and therefore preserve a closer temporal and semantic alignment between the input modality and the endpoint than ICD-code-based labels.
\cite{Soenksen2022mimiciv} also establish unimodal baselines and use Shapley values to quantify the contribution of each modality to the multimodal prediction tasks. This design enables task-specific assessment of modality contribution and provides a useful control for modality-endpoint alignment.

\subsection{Population Alignment, Missingness, and Imputation} \label{sec:imputation}

\begin{figure}[t]
\includegraphics[width=\linewidth]{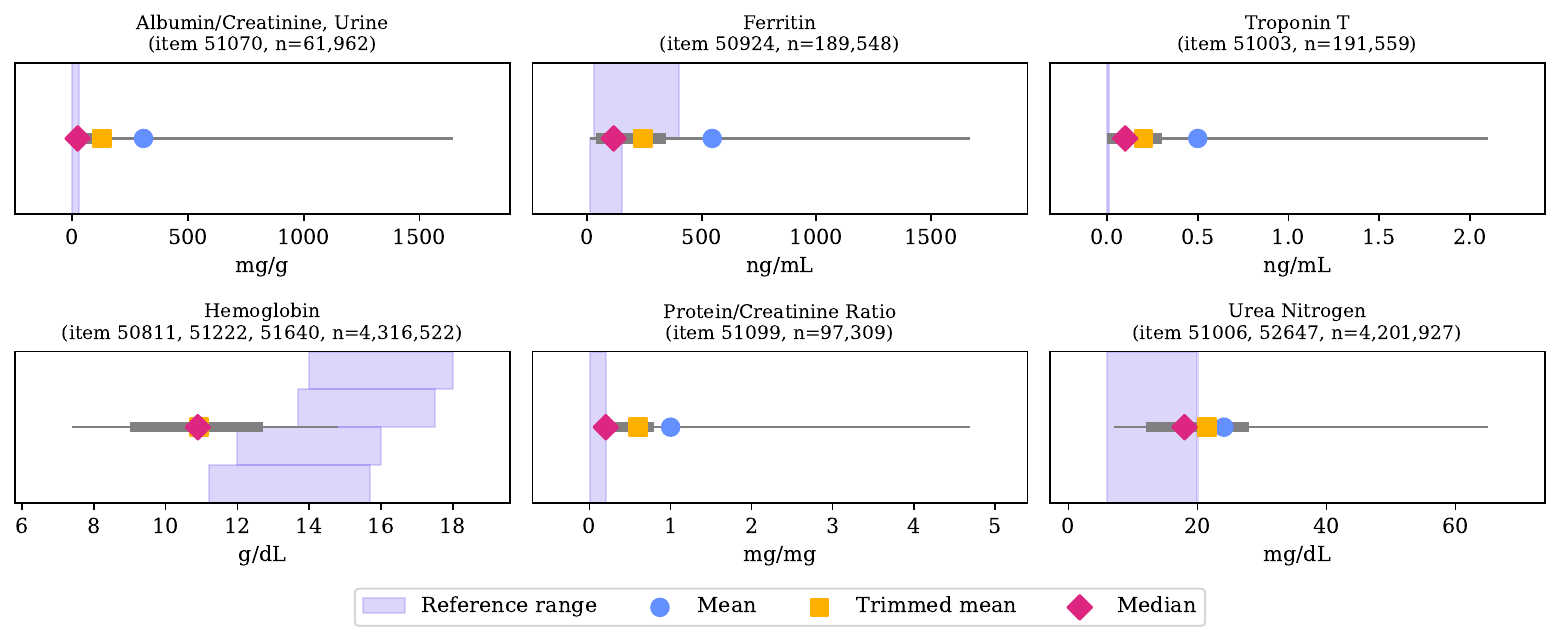}
    \caption{Visualization of descriptive statistics for six laboratory concepts from MIMIC-IV 3.1 ~\citep{PhysioNetMimiciv3.1}. Their names, IDs, units of measurement, and reference ranges are directly taken from MIMIC. We show the reference ranges for each laboratory concept as shaded areas. The 25th to 75th percentile is indicated by a thick line, and the 5th to 95th percentile by a thin line. The mean is denoted as a dot, and the median as a diamond. Since no outliers are excluded and the mean is sensitive to them, we also report the trimmed mean calculated only from values between the 5th and 95th percentiles, visualized as a square.}
    \label{fig:laboratory-stat}
\end{figure}

While previous studies ~\citep{Boag2022EHR, Ghassemi2020EHR, Sauer2022EHRPitfalls} have shown that missingness in EHR data often reflects clinical judgment and other systematic influences, such as human bias or diurnal patterns, the practical implications of this observation for preprocessing have not been explored in sufficient detail. Since not all models can directly handle missing values or irregularly measured time points, preprocessing MIMIC time series data typically involves temporal binning followed by imputation of empty bins. In MIMIC-III, several established preprocessing pipelines ~\citep{Purushotham2018mimic, Liang2021multibench, Wang2020MIMICExtract} employ mean imputation for samples in which measurements are entirely missing. This strategy has been adopted in preprocessing pipelines for MIMIC-IV, effectively replicating the mean-imputation approach used for its predecessor ~\citep{Gupta2022AnEDA, Meng2022DLInterpret, woznica2022consolidated, Zheng2025RealTimePred}. 
Because measurements are ordered selectively in routine care, the distribution of observed values reflects the clinical contexts in which they are measured rather than the full patient population.
As a result, the empirical mean may not correspond to a clinically normal or unobtrusive value or to the empirical mean of the unmeasured patient cohort, leading to \textbf{selection/imputation bias}. Moreover, factors such as age and biological sex can influence what constitutes a normal measurement for an individual ~\citep{Zhernakova2022biomarker}. This issue is demonstrated in Figure \ref{fig:laboratory-stat}, which presents descriptive statistics for selected laboratory concepts whose mean values strongly deviate from the reference ranges provided by MIMIC. Since the recorded minimum and maximum values are often clinically implausible and likely reflect documentation artifacts, we additionally report the mean of values between the 5th and 95th percentiles to reduce the influence of extreme outliers. Nevertheless, this adjusted mean also remains outside the reference range for several concepts. 

This pattern can be explained clinically. For all laboratory values except hemoglobin and ferritin, the 75th and 95th percentiles remain elevated above the reference range, indicating that the documented values cluster in clinically abnormal cases. In routine clinical data, this is expected because these cases are frequently monitored. The albumin-to-creatinine ratio, or more broadly, the protein-to-creatinine ratio in urine, and the urea nitrogen concentration in blood serve as important indicators of renal dysfunction, among other conditions. Accordingly, these measurements exceed their reference ranges in cases of proteinuria ~\citep{Sharma2021Protein1, Kaminska2020Protein2} or elevated urea nitrogen levels ~\citep{Arihan2018Urea, Salazar2014Urea1}. Similarly, Troponin T, a marker of cardiac muscle injury such as myocardial infarction ~\citep{Twerenbold2012Troponin}, is elevated above the reference range in patients with cardiac injury. Ferritin and hemoglobin require a more nuanced interpretation because their reference ranges differ by biological sex. Ferritin reflects iron stores, while hemoglobin reflects the concentration of oxygen-carrying protein in red blood cells. For ferritin, the adjusted mean lies within the adult male reference range but exceeds the upper limit of the female reference range ~\citep{Cullis2018ferritin}. For hemoglobin, MIMIC-IV provides four distinct reference ranges, and the adjusted mean falls below all of them. This is clinically plausible, as illness, blood loss, or major surgical procedures can reduce hemoglobin concentration ~\citep{Karkouti2008Hemo}. These examples show that mean imputation can both erase information carried by missingness and introduce clinically non-neutral values. 

Alternative strategies for time series imputation include median imputation~\citep{Gao2025Mortality, Tang2020FIDDLE}, which is more robust to outliers than mean imputation, as shown in Figure \ref{fig:laboratory-stat}. However, it remains affected by the underlying data distribution. Accordingly, imputed values should be assessed for medical plausibility, regardless of the imputation method used. Because ICU patients differ systematically from healthy cohorts, clinically unremarkable values may not necessarily fall within standard reference ranges; however, this is marker-dependent. The reference ranges provided with MIMIC nevertheless remain informative: when they are stratified by biologically distinct subgroups, such as sex, imputed values should likewise be estimated separately for each subgroup.

In some cases, a clinically motivated static value may be preferable to a data-derived average. For instance, \cite{Gao2025Mortality} impute the fraction of inspired oxygen, defined as the proportion of oxygen in inhaled air, with 21\%, corresponding to atmospheric air, when no recorded measurement indicates that the patient is receiving oxygen-enriched air. This is clinically reasonable because the fraction of inspired oxygen (item IDs: 226754, 227010, 229280, 229841) is most frequently documented only when patients require supplemental oxygen. In contrast, imputing the mean (88.5\%), adjusted mean (91.4\%), or median (100.0\%) would be misleading for a patient with entirely missing values who is breathing room air. Thus, assigning a constant value \citep{Gao2025Mortality, Harutyunyan2019MIMICMultiTask} may be more clinically plausible and may help reduce bias introduced by imputation, although any imputed value will differ from the ground truth and may therefore carry bias. Accordingly, the selected values should be reported to ensure transparency and reproducibility. 

As any form of imputation erases the implicit information of clinical judgment of patient state in missingness itself, missingness indicators are a useful tool for preserving this information. These auxiliary binary variables ~\citep{Sharafoddini2019missingnessindicator, Wang2020MIMICExtract, Harutyunyan2019MIMICMultiTask, Tang2020FIDDLE} encode the presence or absence of measurements and are added as additional input variables. For aggregated features, the percentage of missingness can be considered instead.

A related but distinct issue of \textbf{selection/imputation bias} is that several modules in MIMIC-IV cover only subsets of the full cohort. For instance, only 47.8\% of emergency department stays result in a hospital admission, while only 37.1\% of admissions have a linked emergency department stay. Furthermore, microbiology data is only available for 36.8\% of admissions, discharge notes for 60.8\%, ICU stays for 15.6\%, and radiology notes for 56.7\%. These coverage gaps reflect systematic factors, such as module-specific data collection periods and heterogeneity in patient care pathways. When constructing feature matrices from multiple modules, using the full MIMIC-IV admission cohort as the analytic population can therefore introduce structural missingness that is not directly attributable to clinical judgment or patient state. 

Imputing values from the statistical properties of an observed subgroup onto the broader cohort is inadvisable because the observed subgroup generally cannot be assumed to represent the full population. Clear subgroup differences, such as the length of stay, further support this concern. Admissions with laboratory documentation had a median length of stay of 3.6 days, compared with 0.7 days for admissions without such documentation. For microbiology data, admissions with available documentation had a median length of stay of 5.4 days, compared with 1.8 days for admissions without microbiology documentation. Similarly, the in-hospital mortality rate was 11.1\% among admissions with ICU stays, compared with only 0.5\% among admissions without documented ICU stays. These differences indicate that module availability often reflects fundamentally different patient populations and care trajectories.

Consequently, unless a model architecture is explicitly designed to accommodate missing modalities, the more principled approach is to exclude either the affected features or the affected admissions, rather than to impute across a subgroup whose representativeness cannot be established.

\section{Empirical Case Studies}
\begin{table}[!h]
\centering
\caption{Time series and tabular variables included in the dataset.}\label{tbl-exp-variables}
\resizebox{0.8\textwidth}{!}{
\begin{tabular}{lllll}
\toprule
Name&Modality&Type&Unit&Static value\\
\midrule
\textbf{Demographics}&&&&\\
\rowcolor{gray!8}Age & tabular & discrete&years&- \\
 Weight & tabular & continuous& kg & median\\
\rowcolor{gray!8}Sex& tabular & nominal& -&-\\
Ethnicity & tabular & nominal&-&-\\
\rowcolor{gray!8}Height& tabular & continuous& cm & median\\
\midrule
\textbf{Laboratory values} &&&& \\
\rowcolor{gray!8} Bicarbonate & time series & continuous &mEq/L& 27.0\\
   Bilirubin & time series & continuous &mg/dL& 0.7\\
 \rowcolor{gray!8} Blood urea nitrogen & time series & continuous&mg/dL& 14.0 \\
 Chloride & time series & continuous&mEq/L & 100.0\\
\rowcolor{gray!8}  Creatinine & time series & continuous& mg/dL & 1.0\\
   Glucose & time series & continuous&mg/dL & 85.0 \\
  \rowcolor{gray!8} Potassium & time series & continuous& mEq/L & 4.0\\
   pH & time series & continuous&- & 7.4 \\
\midrule
\textbf{Clinical scores}&&&\\
\rowcolor{gray!8} Glasgow Coma Scale eye opening & time series & ordinal&- &4\\
  Glasgow Coma Scale motor response & time series & ordinal&-&6 \\
\rowcolor{gray!8} Glasgow Coma Scale verbal response  & time series & ordinal&-&5 \\
 \midrule
\textbf{Vital parameters} &&&&  \\
\rowcolor{gray!8} Diastolic blood pressure & time series & continuous&mmHg& 80.0 \\
 Fraction inspired oxygen& time series & continuous& fraction & 0.21 \\
 \rowcolor{gray!8} Heart Rate& time series & continuous& bpm & 80.0\\
  Mean blood pressure  & time series & continuous& mmHg & 90.0\\
  \rowcolor{gray!8}Oxygen saturation& time series & continuous& \% & 99.0 \\
  Respiratory rate& time series & continuous& breaths/min &  15.0\\
   \rowcolor{gray!8}Systolic blood pressure & time series & continuous&mmHg & 120.0\\
  Temperature & time series & continuous& Celsius & 36.6\\
\bottomrule
\end{tabular}
}
\end{table}

To evaluate the practical influence of certain potential preprocessing pitfalls within the EHR-data pipelines outlined above, we devised experiments centered on two prevalent prediction tasks: 25 phenotype-specific binary prediction tasks and in-hospital mortality prediction from the first 48 hours of an ICU stay. Specifically, we examine the effects of various preprocessing steps on in-hospital mortality prediction and address the potential modality-endpoint mismatch in the 25 phenotype prediction tasks by incorporating images as an additional modality, reducing the cohort for this experiment to stays that have imaging data available. We adapted an existing preprocessing pipeline ~\citep{hayat2023medfuse, Rasekh2024Fusion} and drew inspiration from \cite{Soenksen2022mimiciv} regarding feature engineering, model selection, and feature importance.
All train, validation, and test partitions are constructed at the patient level using \texttt{subject\_id}, such that all admissions and ICU stays belonging to the same patient remain within a single partition. Patients with less than 48 hours of recorded data during their ICU stay are excluded from the dataset. For the phenotype experiment, the cohort is reduced to only stays with at least one anteroposterior chest X-ray available during the first 48 hours. Appendix~\ref {app:data-description} Tables~\ref {tbl-data-description-mort} and ~\ref{tbl-data-description-pheno} report cohort sizes, demographic characteristics and outcome prevalence for the mortality and phenotype experiments, respectively.

A primary preprocessing decision is the choice of timestamp for chart events. We extract two versions of the dataset: one using \texttt{charttime}, corresponding more closely to an event-time or patient-state perspective, and another using \texttt{storetime}, which provides a more conservative proxy for information availability under a clinician-knowledge interpretation. For all included laboratory concepts, we only select item IDs that match in both unit and specimen fluid. Table~\ref{tbl-exp-variables} lists all chosen tabular and time series variables. 
For both dataset versions, we aggregate raw time series into hourly bins over the 48-hour prediction window. Bins containing at least one real observation are filled using a forward-then-backward fill strategy. For variables without an observation within the 48-hour window, we implement two imputation strategies: (1) global training-set mean/mode imputation, observed as the default statistical imputation strategy in many pipelines (see Section~\ref{sec:imputation}), and (2) clinically plausible static value imputation, serving as a more robust alternative. The chosen values are reported in Table~\ref{tbl-exp-variables}. From the resulting dataset, we compute the same eleven features per time series variable used by HAIM ~\citep{Soenksen2022mimiciv}. Additionally, we optionally add a twelfth feature indicating the number of hourly bins containing an actual measurement to capture missingness information. For tabular data, we apply (1) mean or (2) median imputation for missing continuous values (height and weight), with an optional binary missingness variable encoding whether the value is real or imputed. For the chest X-rays, we select the latest anteroposterior chest view study taken within the first 48 hours. We normalize each image, center-crop it, resize it to $224\times224$ pixels, and, following \cite{Soenksen2022mimiciv}, extract 1024-dim dense embeddings as features using a DenseNet121 pretrained on CheXpert.

For the mortality endpoint, we compare the two imputation regimes across both timestamp versions. We refer to the mean/mode imputation as \textit{statistical} preprocessing, and the clinically plausible imputation as \textit{static value} preprocessing. Additionally, we evaluate each imputation method and timestamp choice both without and with missingness features; the latter is indicated by the additional label \textit{+ missing}. For the phenotype task, we utilize only the \textit{static value} preprocessing strategy and limit the cohort to stays with available imaging. This allows us to evaluate both the overall importance attributed to image-modality features and the impact of the image modality on the performance of all binary phenotype prediction models. We follow \cite{Soenksen2022mimiciv} in using XGBoost for all variants. All models use an identical hyperparameter configuration (350 estimators, maximum depth 4, learning rate 0.05, row/column subsampling of 0.8). We repeat each experiment using five fixed random seeds and average the results.

\begin{table}[!h]
\centering
\caption{Averaged results over 5 runs of predicting in-hospital mortality with models trained on dataset variations of timestamp choice (chart- or storetime), imputation strategy  (statistical global mean/mode or static value imputation), and optional inclusion of missingness information.}\label{tbl-exp-results}
\resizebox{0.8\linewidth}{!}{
\begin{tabular}{lcc}
\toprule
Preprocessing Approach& \multicolumn{2}{c}{In-hospital Mortality Prediction} \\
\cmidrule(lr){2-3}
 & AUROC in \% & AUPRC in \%  \\
\midrule
 \rowcolor{gray!8} Static value storetime&$89.34_{\pm0.10}$&$59.07_{\pm0.40}$\\
  Statistical storetime&$89.28_{\pm0.09}$&$58.57_{\pm0.42}$\\
 \rowcolor{gray!8} Static value charttime &$89.41_{\pm0.06}$&$59.71_{\pm0.44}$\\
 Statistical charttime&$89.36_{\pm0.12}$&$59.29_{\pm0.70}$\\
  \rowcolor{gray!8} Static value storetime + missing &$89.70_{\pm0.09}$&$60.44_{\pm0.19}$\\
  Statistical storetime + missing &$89.56_{\pm0.09}$&$60.17_{\pm0.39}$\\
 \rowcolor{gray!8} Static value charttime + missing&$89.69_{\pm0.01}$&$60.79_{\pm0.16}$\\
 Statistical charttime + missing&$89.64_{\pm0.08}$&$60.40_{\pm0.37}$\\
\bottomrule
\end{tabular}
}
\end{table}

\begin{figure*}[!h]
    \centering
    \includegraphics[height=0.81\textheight]{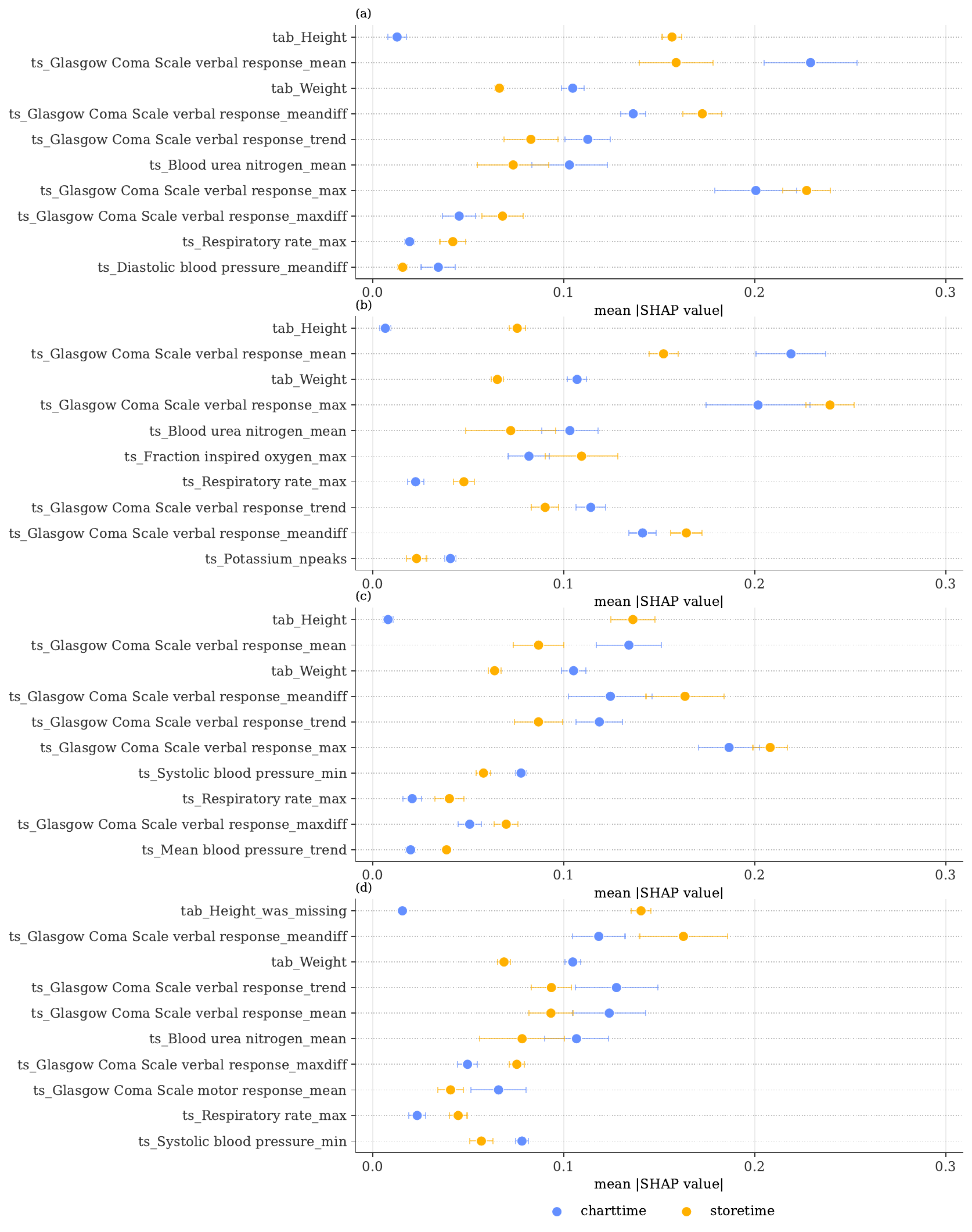}
    \caption{Differences between mean SHAP values across charttime and storetime timestamps for predicting in-hospital mortality in models trained with data preprocessed using: (a) statistical mean/mode imputation, (b) static value imputation, (c) statistical mean/mode imputation with missingness information, and (d) static value imputation with missingness information.}
    \label{fig:shap_comparison}
\end{figure*}

\subsection{In-hospital Mortality}
For this experiment, we adopt a clinician-knowledge interpretation at the 48-hour prediction anchor: the model is evaluated as if it were making a prediction using information that could reasonably have been available by that time. Accordingly, the \texttt{storetime} variant is interpreted as the more conservative availability-aware representation, whereas the \texttt{charttime} variant provides a patient-state-oriented comparison rather than being treated as intrinsically invalid preprocessing. Table~\ref{tbl-data-description-mort} in Appendix \ref{app:data-description} shows differences in missingness across timestamps. For laboratory values, complete missingness increases slightly when \texttt{storetime} is used. In contrast, partial missingness increases sharply for some vital parameters when \texttt{storetime} is applied, indicating that manual input or clinical staff verification occurs less frequently, creating gaps in the time-series data.
The performance results of all eight tested model configurations are in Table \ref{tbl-exp-results}. They show a slight numerical improvement in AUROC and AUPRC for both timestamp versions, with and without missingness, when using the static value preprocessing strategy compared to the statistical preprocessing. Notably, \texttt{storetime} also performs marginally worse than the corresponding \texttt{charttime} version in AUPRC across all settings and in AUROC in three of four. Adding missingness also has a small numerical improvement on each timestamp and imputation method. Overall, all eight experiments yield similar results, with less than a 0.5\% difference in AUROC and roughly a 2.22\% difference in AUPRC. While different preprocessing and timestamp choices appear to have only a minimal effect on aggregate performance, Figure \ref{fig:shap_comparison} shows this is not an argument against careful preprocessing. Here, the features where \texttt{charttime} and \texttt{storetime} models differ most substantially in mean overall feature importance are presented, highlighting distinct differences in feature importance. 
Comparisons (a) to (c) indicate that the \texttt{storetime} models rely substantially more on body height than the \texttt{charttime} model, particularly when \textit{statistical} imputation is applied. This could suggest an association between body height and survival when a \texttt{storetime}-based prediction model is used for data analysis. However, Figure~\ref{fig:shap_comparison}(d), \textit{static value + missing}, indicates that the signal is associated not with height itself, but with its documentation status. The imputation process itself provides a detectable signal of missingness used by the \textit{storetime} models, particularly when mean imputation is applied. Conversely, when missingness information is incorporated, and the median is used for imputation, as shown in Figure~\ref{fig:shap_comparison}(d), it becomes clear that the model exploits documentation differences that likely arise from differences in patient status and care pathways, which affect documentation practices. Specifically, mortality is approximately 4.5\% higher in the group of patients with documented height than in those without. The stark difference between the importance of the height documentation status for \texttt{storetime} models compared to \texttt{charttime} models could stem from the higher number of time series measurements, and therefore information signal, available to \texttt{charttime}.

Overall, the results demonstrate that preprocessing errors can go unnoticed as they do not necessarily manifest in performance metrics in obvious ways. However, as visible in the SHAP feature importance analysis, different preprocessing choices can alter the features the models rely on and, consequently, the logic underlying the predictions.

\begin{figure}[!h]
    \centering
    \includegraphics[width=\linewidth]{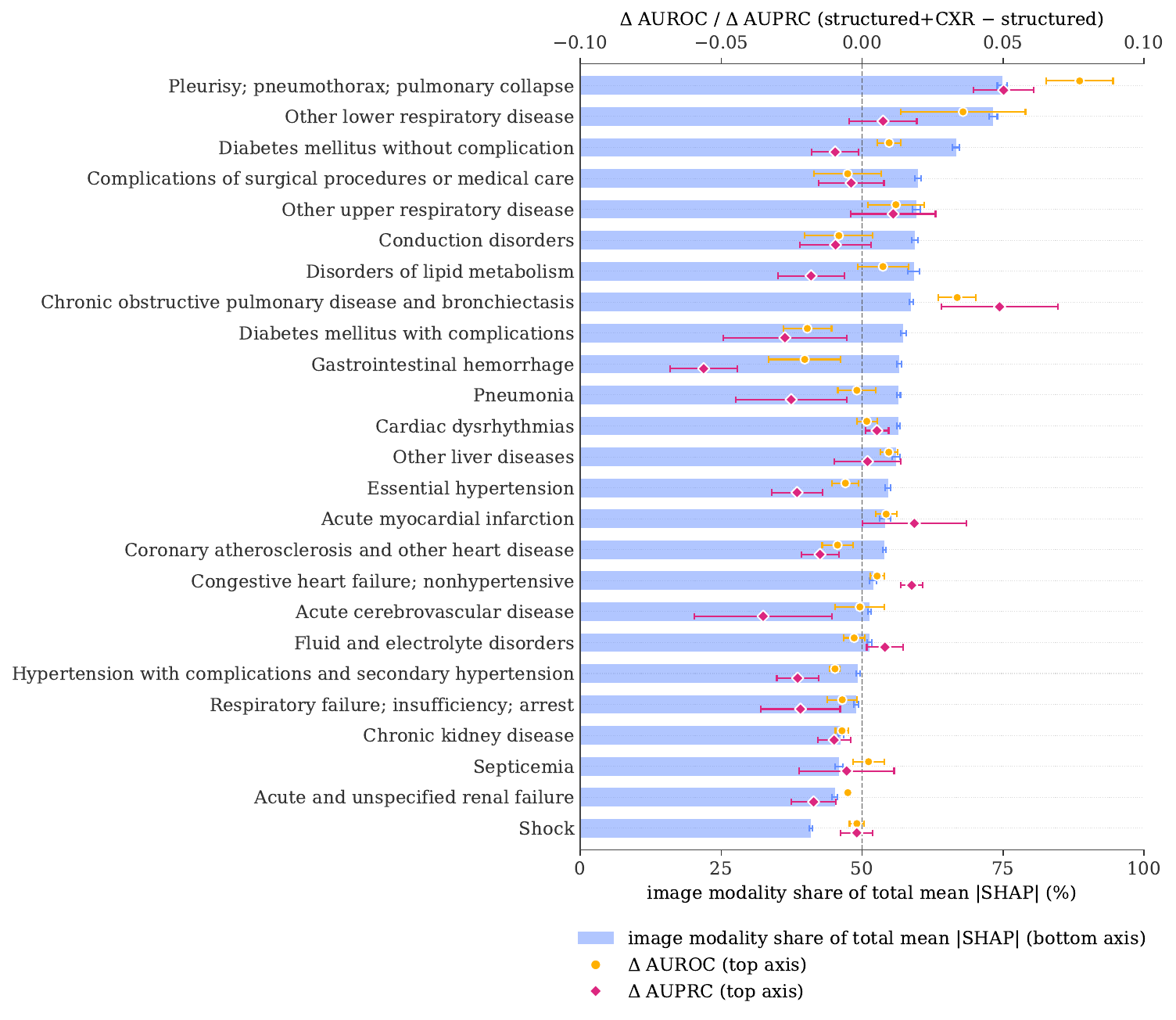}
    \caption{The percentage of total mean SHAP values assigned to features derived from the image modality for each phenotype prediction task (lower x-axis) and the performance delta between only structured data and structured data with CXR as input features (upper x-axis). Model performance delta for each phenotype is reported for AUROC and AUPRC. Mean values and confidence intervals are reported over five runs.}
    \label{fig:shap_image_pheno}
\end{figure}

\subsection{Phenotype Classification}
For this experiment, we again adopt a clinician-knowledge interpretation at the 48-hour prediction anchor. Accordingly, the more conservative \texttt{storetime} timestamp is used. 
Figure~\ref{fig:shap_image_pheno} presents both the overall mean SHAP values assigned to features in the image modality and the changes in AUROC and AUPRC for each phenotype resulting from the inclusion of the image modality. The observable discrepancy between large aggregate SHAP attribution to the image representation (40.9\% to 74.8\% of the overall mean importance) and the limited—and, in some cases, negative—improvement in predictive performance indicates that aggregate feature attribution alone should not be interpreted as evidence of clinically meaningful modality contribution. The high dimensionality of the image representation is one possible explanation and warrants dedicated control experiments. Thus, high aggregate image feature importance does not necessarily imply that the imaging modality improves prediction or contributes endpoint-relevant signal.

\section{Discussion, Limitations, and Conclusions}
Using MIMIC-IV, we show that violations of temporal, semantic, and population alignment can alter the information presented to a model and the features it relies on, even when aggregate predictive performance changes only modestly. 
To address these challenges and mitigate their associated risks, we propose the following recommendations for designing and evaluating MIMIC preprocessing pipelines:
\begin{enumerate}[nosep] 
    \item \textbf{Prioritizing transparency:} Document all variable selection criteria, general preprocessing decisions, timestamp choices, outlier handling procedures, and imputation strategies to support reproducibility. Explicitly address potential leakage, bias, or limitations introduced by the dataset structure and preprocessing pipeline.
    \item \textbf{Ensuring temporal validity:} Define the modeling perspective and prediction anchor explicitly, and align each feature using the timestamp that corresponds to the information state being modeled. Report source tables, timestamp fields, and availability assumptions.
    \item \textbf{Ensuring endpoint and feature-task validity:} Define endpoints precisely, avoid conflating clinically distinct phenomena, and verify that selected features and modalities can plausibly provide evidence for the target at the intended prediction time.
    \item \textbf{Applying context-aware imputation:} Select imputation strategies according to the variable and missingness mechanism, assess the plausibility of imputed values, and preserve informative missingness explicitly where appropriate.
   
    \item \textbf{Avoid unjustified imputation or exclusion:} When missingness makes reliable imputation impossible, consider excluding affected variables or episodes, but report the resulting attrition and assess whether exclusion changes the represented population.
    \item \textbf{Validate multimodal contributions:} For multimodal tasks, compare multimodal models against matched unimodal and modality-ablation baselines. Attribution methods such as SHAP, as proposed by e.g., \cite{Soenksen2022mimiciv}, can provide complementary information about model reliance, but should not be interpreted alone as evidence that a modality provides clinically meaningful or endpoint-relevant information.
\end{enumerate}
A more extensive checklist is provided in Appendix \ref{app:checklist} to guide and support the development and validation of preprocessing pipelines. The audit is not exhaustive, and additional dataset- or task-specific failure modes may require further safeguards. Therefore, ongoing vigilance and careful consideration of preprocessing decisions remain essential. Because many of these failure modes arise from routine clinical documentation and workflow, they are likely relevant to EHR datasets beyond MIMIC. 

\textbf{Limitations.} Our analysis is based on MIMIC-IV, which originates from a single healthcare institution, and some timestamp semantics, documentation practices, and workflow conventions may therefore be institution- or system-specific. \texttt{storetime} should be interpreted as an available proxy for information availability rather than a direct record of what an individual clinician had observed at a given moment. The clinically informed imputation values examined here are heuristics rather than ground-truth values and may require adaptation to other populations and tasks. SHAP values characterize model attribution but do not establish causal or clinically meaningful modality contribution and can be affected by feature dimensionality and correlation. Furthermore, excluding patients, variables, or modalities because of missingness can itself introduce selection bias. Finally, the literature examples discussed in Section~\ref{sec:prep} are illustrative assessments based on publicly reported methods and implementations rather than a systematic review, and external validation in other EHR systems remains necessary.

\section*{Data Availability}
The public dataset MIMIC-IV 3.1 ~\citep{PhysioNetMimiciv3.1}, as well as its extending modules MIMIC-IV-Note 2.2 ~\citep{PhysioNet-mimic-iv-note-2.2}, MIMIC-CXR 2.1.0 ~\citep{Johnson2019MIMICCXR, PhysioNet-mimic-cxr-jpg-2.1.0}, MIMIC-IV-ECG 1.0 ~\citep{PhysioNet-mimic-iv-ecg-1.0} and MIMIC-IV-ED 2.2 ~\citep{PhysioNet-mimic-iv-ed-2.2} are available from PhysioNet ~\citep{PhysioNet} with the required credentials.

\impact{Our contribution addresses several preprocessing issues that may negatively affect the results, interpretation, and understanding of research involving EHR data. This work may raise awareness of these issues and, in many cases, inform readers of better-suited alternatives in these scenarios. The checklist derived from the paper also provides a practical resource to guide the selection, verification, or development of a pipeline. A potential negative impact is that the checklist could be interpreted as exhaustive. It should instead complement dataset-specific clinical and technical validation.
}


\acks{This work was supported by the Federal Ministry of Research, Technology and Space of Germany [project name: EMuLe- Enhancing Data and Model Efficiency in Multimodal Learning; grant number 16IS24059]. The third and last authors were partially funded by the Lower Saxony Ministry of Science and Culture (MWK) with funds from the Volkswagen Foundation’s Zukunft Niedersachsen program [project name: CAIMed- Lower Saxony Center for Artificial Intelligence and Causal Methods in Medicine; grant number: ZN4257]. The second author was funded by the European Union under the grant project 101079894 (COMFORT - Improving Urologic Cancer Care with Artificial Intelligence Solutions). Views and opinions expressed are, however, those of the author(s) only and do not necessarily reflect those of the European Union or the European Health and Digital Executive Agency (HADEA). Neither the European Union nor the granting authority can be held responsible for them. 

The authors acknowledge Hannover Medical School for providing MHH-HPC resources that have contributed to the findings reported in this paper. The authors appreciate AI Grid, Germany’s network for young AI researchers, funded by the Federal Ministry of Research, Technology, and Space (BMFTR), for fostering the scientific exchange from which this collaboration emerged.

The authors declare that they have no known competing financial interests or personal relationships that could have appeared to influence the work reported in this paper.
}

\vskip 0.2in
\bibliography{cas-refs}

\appendix
\section{}\label{app:mimic_schema}
\begin{figure*}[h]
    \centering
    \includegraphics[width=\linewidth]{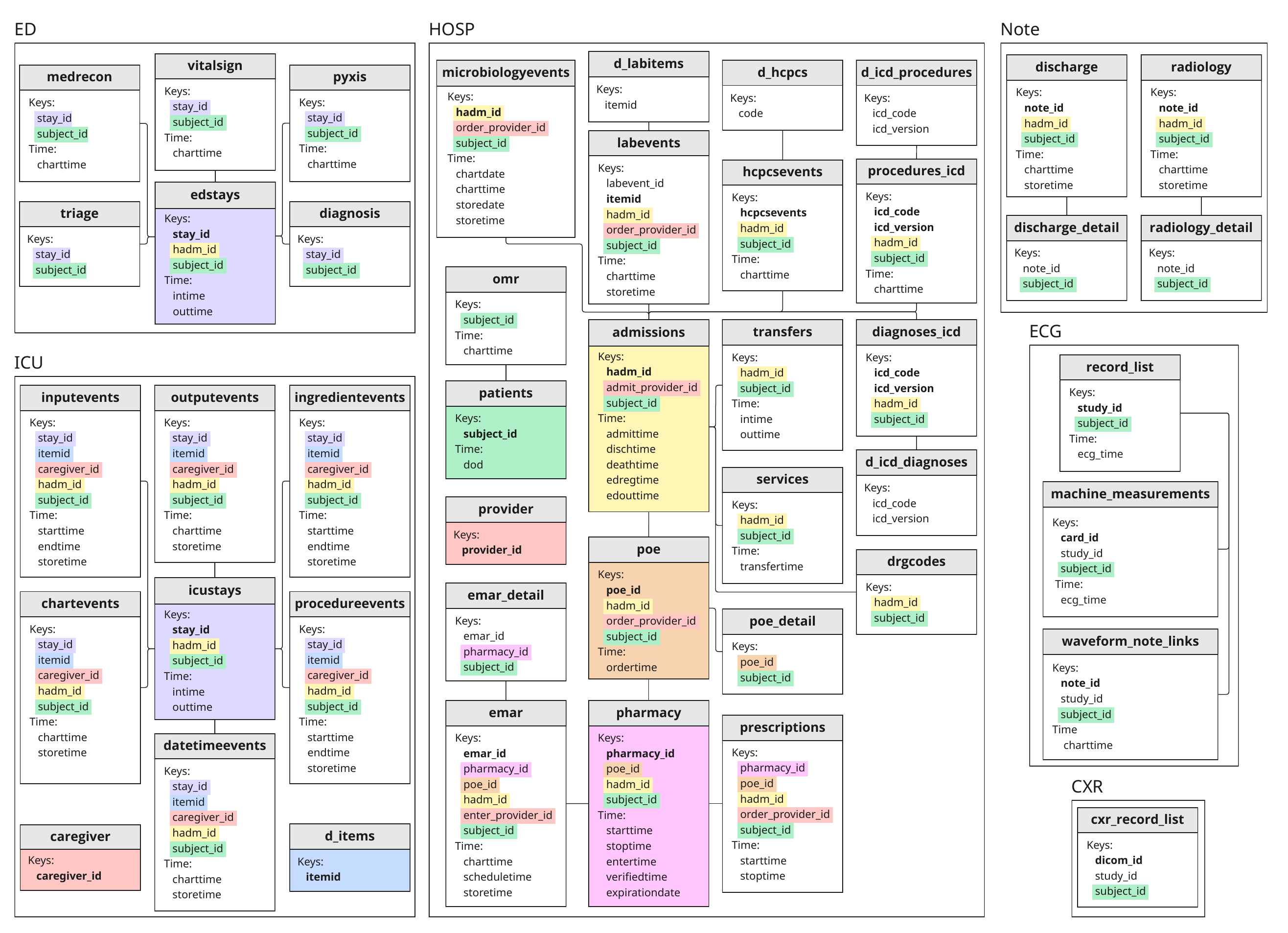}
    \caption{Overview of the MIMIC-IV modules and tables included in the schema. The figure emphasizes linkage identifiers and temporal fields that are relevant for episode alignment and the resolution of temporal ambiguity rather than listing every available variable. Colors indicate recurring identifier scopes across modules, including the global patient identifier (\texttt{subject\_id}), encounter- and stay-level identifiers (\texttt{hadm\_id}, \texttt{stay\_id}), and subsystem-specific local identifiers. Tables containing the complete list of variables for all included MIMIC-IV tables are provided in the Appendix. For the complete official schema, including table and column descriptions, see the MIMIC-IV documentation at \url{https://mimic.mit.edu}.}
    \label{fig:mimiciv_schema}
\end{figure*}

\begin{figure*}[h]
    \centering
    \includegraphics[height=0.9\textheight]{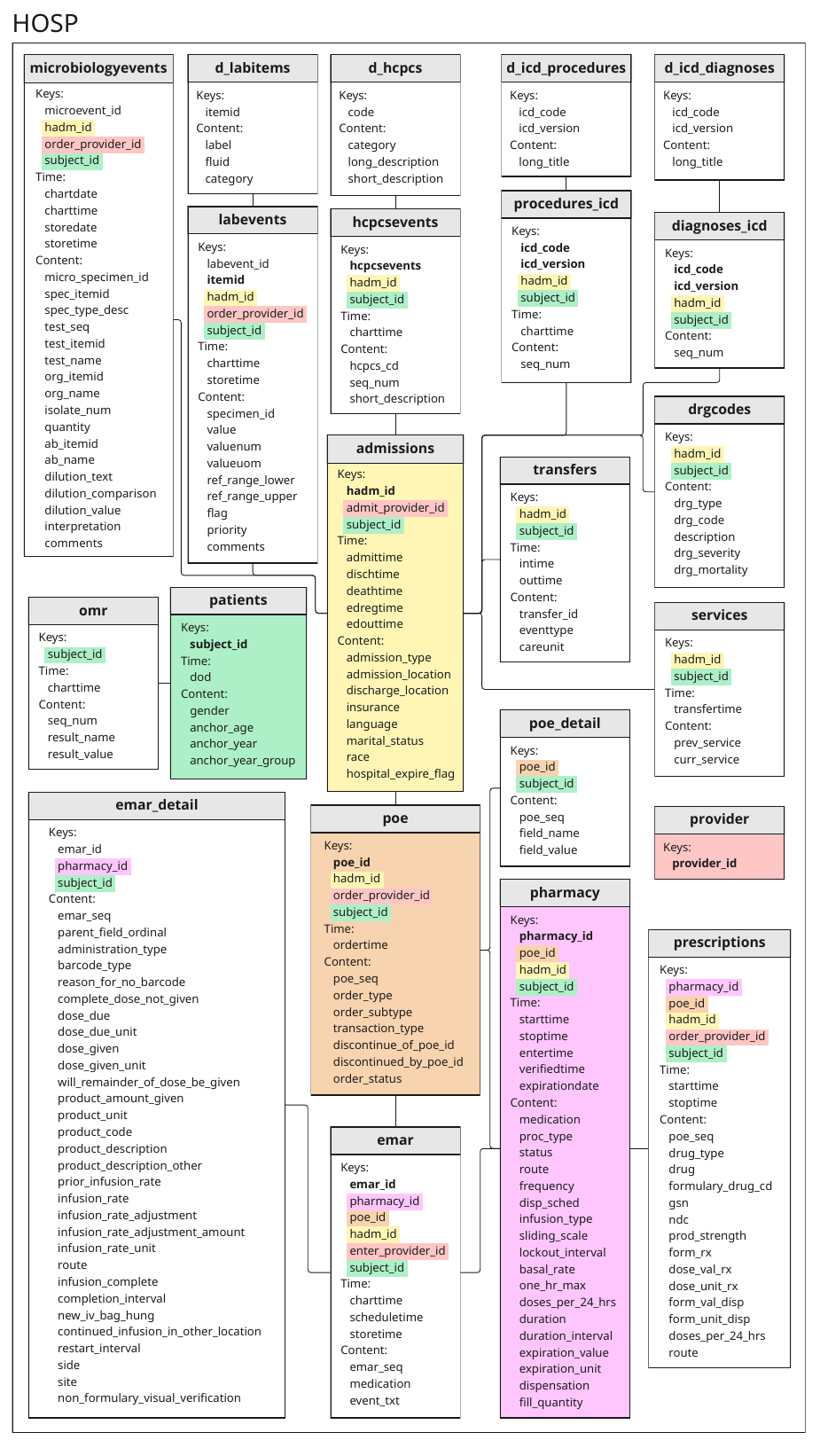}
    \caption{Schema overview of the MIMIC-IV \texttt{hosp} module, highlighting the main tables, linkage identifiers, and temporal fields.}
    \label{fig:mimiciv_schema_ext1}
\end{figure*}

\begin{figure*}[h]
    \centering
    \includegraphics[height=0.9\textheight]{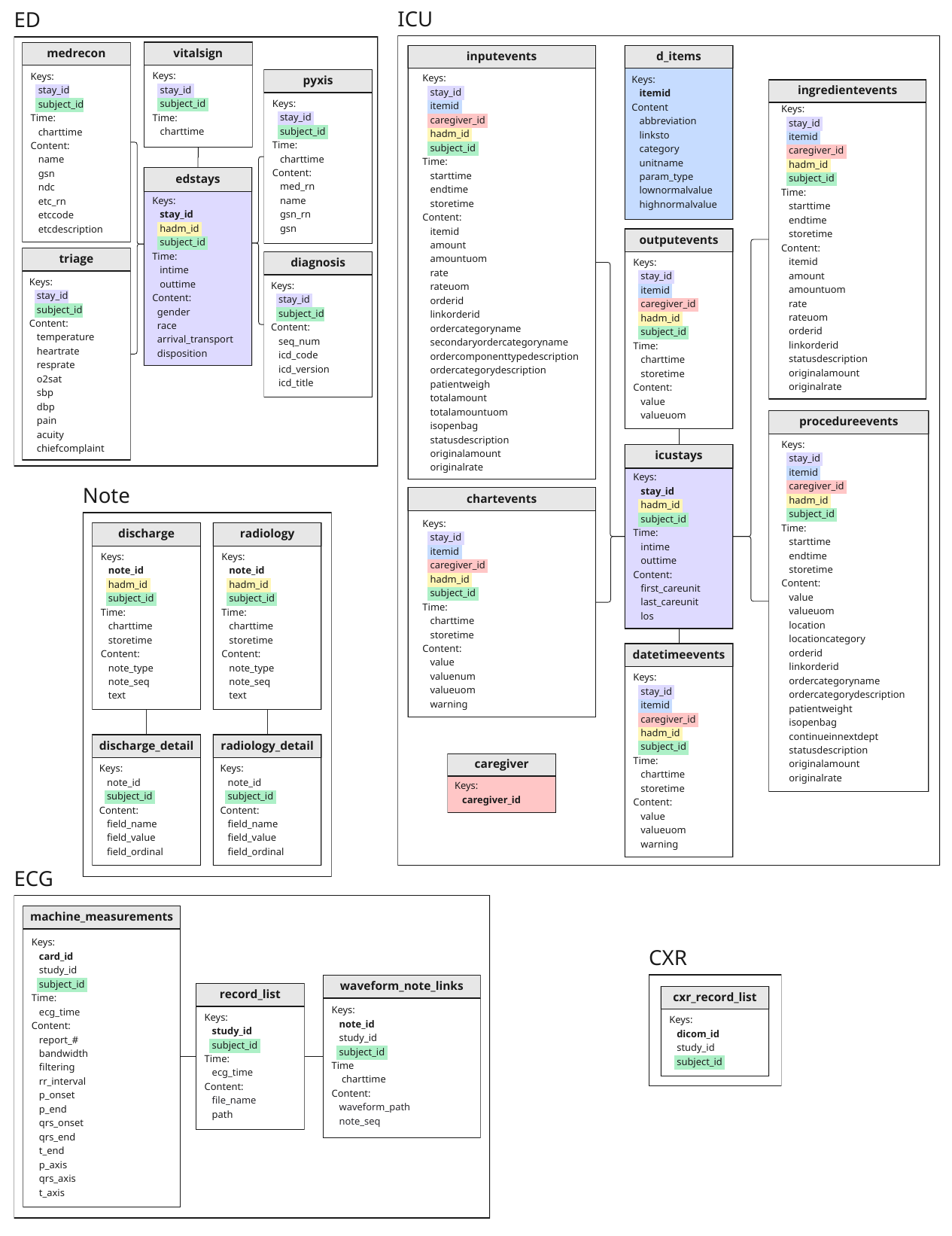}
    \caption{Schema overview of the MIMIC-IV ED, ICU, note, ECG, and CXR modules, highlighting the main tables, linkage identifiers, and temporal fields.}
    \label{fig:mimiciv_schema_ext2}
\end{figure*}
\clearpage
\section{}\label{app:data-description}
\begin{table}
\centering
\caption{Cohort description for the in-hospital mortality experiment by data split. For each time series variable, partial measurements (at least one bin measured, but not all) and completely missing measurements (no bin measured) within the 48h window are given. For tabular data, only complete missingness is reported. Chart- and storetime variants are listed separately.}\label{tbl-data-description-mort}
\resizebox{0.6\linewidth}{!}{
\begin{tabular}{llll}
\toprule
 &Train&Validation&Test\\
\midrule
\rowcolor{gray!8}Unique patients, n & 25,276 & 1,758&4,411 \\
 Unique stays, n & 29,171 & 2,161& 5,302 \\
\rowcolor{gray!8}Age, mean& $63.76_{\pm16.37}$ & $64.41_{\pm16.01}$& $63.95_{\pm16.37}$\\
\textbf{Sex, \%} &  & &\\
\rowcolor{gray!8}Female& 44.08 & 43.54& 43.29 \\
 Male & 55.92 & 56.46 & 56.71\\
 \rowcolor{gray!8}\textbf{Ethnicity, \%} &  & &\\
White& 65.13 & 68.53& 67.92 \\
 \rowcolor{gray!8}Other/Unknown & 17.50 & 13.93 &14.77\\
 Black& 10.65&9.90&10.9\\
 \rowcolor{gray!8}Hispanic&3.78&3.75&3.70\\
 Asian&2.94&3.89&2.72\\
\rowcolor{gray!8} In-hospital mortality, \%& 13.65& 12.36&12.58\\
\textbf{Chart: partial/complete missing, \%}&&&\\
\rowcolor{gray!8}Bicarbonate&98.87/1.13&98.70/1.30&99.08/0.92\\
Bilirubin&55.66/44.34&52.11/47.89&52.24/47.76\\
\rowcolor{gray!8}Blood urea nitrogen&99.41/0.59&99.54/0.46&99.68/0.32\\
Chloride&99.15/0.85&99.40/0.60&99.53/0.47\\
\rowcolor{gray!8}Creatinine&99.08/0.92&98.98/1.02&99.21/0.79\\
Diastolic blood pressure&79.15/0.17&78.81/0.19&79.39/0.19\\
\rowcolor{gray!8}Fraction inspired oxygen&62.49/37.51&63.17/36.83&62.83/37.17\\
Glasgow coma scale eye opening&96.15/3.75&97.32/2.68&96.66/3.23\\
\rowcolor{gray!8}Glasgow coma scale motor response&99.60/0.17&99.91/0.09&99.60/0.19\\
Glasgow coma scale verbal response&99.59/0.18&99.86/0.14&99.59/0.19\\
\rowcolor{gray!8}Glucose&8.29/91.71&11.57/88.43&10.96/89.04\\
Heart Rate&66.04/0.14&63.77/0.05&63.96/0.17\\
\rowcolor{gray!8}Mean blood pressure&78.61/0.16&77.83/0.05&78.35/0.17\\
Oxygen saturation&71.20/0.07&69.88/0.00&70.22/0.09\\
\rowcolor{gray!8}Potassium&93.68/6.32&93.38/6.62&91.85/8.15\\
Respiratory rate&66.22/0.24&64.69/0.19&64.52/0.21\\
\rowcolor{gray!8}Systolic blood pressure&79.06/0.17&78.67/0.19&79.22/0.17\\
Temperature&98.18/1.70&97.59/2.17&98.11/1.87\\
\rowcolor{gray!8}pH&71.22/28.78&72.98/27.02&72.20/27.80\\
Height&-/39.21&-/37.76&-/39.97\\
\rowcolor{gray!8}Weight&-/0.03&-/0.00&-/0.04\\
\textbf{Store: partial/complete missing, \%}&&&\\
\rowcolor{gray!8}Bicarbonate&98.84/1.16&98.66/1.34&99.06/0.94\\
Bilirubin&55.53/44.47&51.69/48.31&52.11/47.89\\
\rowcolor{gray!8}Blood urea nitrogen&99.40/0.60&99.49/0.51&99.66/0.34\\
Chloride&99.13/0.87&99.35/0.65&99.47/0.53\\
\rowcolor{gray!8}Creatinine&99.08/0.92&98.94/1.06&99.19/0.81\\
Diastolic blood pressure&99.80/0.17&99.72/0.28&99.79/0.19\\
\rowcolor{gray!8}Fraction inspired oxygen&62.51/37.49&63.21/36.79&62.83/37.17\\
Glasgow coma scale eye opening&96.21/3.79&97.27/2.73&96.77/3.23\\
\rowcolor{gray!8}Glasgow coma scale motor response&99.83/0.17&99.86/0.14&99.83/0.17\\
Glasgow coma scale verbal response&99.83/0.17&99.81/0.19&99.83/0.17\\
\rowcolor{gray!8}Glucose&8.28/91.72&11.57/88.43&10.96/89.04\\
Heart Rate&99.83/0.15&99.86/0.14&99.81/0.17\\
\rowcolor{gray!8}Mean blood pressure&99.81/0.16&99.86/0.14&99.81/0.17\\
Oxygen saturation&99.90/0.07&99.86/0.09&99.91/0.09\\
\rowcolor{gray!8}Potassium&93.67/6.33&93.34/6.66&91.85/8.15\\
Respiratory rate&99.73/0.25&99.77/0.23&99.74/0.21\\
\rowcolor{gray!8}Systolic blood pressure&99.80/0.17&99.72/0.28&99.81/0.17\\
Temperature&98.29/1.71&97.78/2.22&98.25/1.75\\
\rowcolor{gray!8}pH&71.26/28.74&72.93/27.07&72.24/27.76\\
Height&-/41.92&-/41.09&-/41.91\\
\rowcolor{gray!8}Weight&-/0.05&-/0.00&-/0.06\\
\bottomrule
\end{tabular}
}
\end{table}

\begin{table}
\centering
\caption{Cohort description for the phenotype prediction experiment separated by train/validation/test split.}\label{tbl-data-description-pheno}
\resizebox{0.8\linewidth}{!}{
\begin{tabular}{llll}
\toprule
 &Train&Validation&Test\\
\midrule
\rowcolor{gray!8}Unique patients, n & 4,449 & 419 & 1,118 \\
 Unique stays, n & 4,956 & 490 & 1,231 \\
\rowcolor{gray!8}Age, mean& $63.72_{\pm16.59}$ & $63.27_{\pm16.40}$& $63.54_{\pm16.35}$\\
\textbf{Sex, \%} &  & &\\
\rowcolor{gray!8}Female& 45.16 & 42.86& 43.38 \\
 Male & 54.84 & 57.14 & 56.62\\
 \rowcolor{gray!8}\textbf{Ethnicity, \%} &  & &\\
White& 65.01 & 66.53& 64.42 \\
 \rowcolor{gray!8}Other/Unkown & 15.27 & 14.49 &15.76\\
 Black& 12.77&10.82&13.00\\
 \rowcolor{gray!8}Hispanic&3.67&4.69&3.98\\
 Asian&3.27&3.47&2.84\\
\textbf{Label positive rate, \%}&&&\\
\rowcolor{gray!8}Acute and unspecified renal failure&38.48&38.37&37.37\\
Acute cerebrovascular disease&9.46&7.55&8.53\\
\rowcolor{gray!8}Acute myocardial infarction&9.77&8.57&9.91\\
Cardiac dysrhythmias&41.10&35.31&40.05\\
\rowcolor{gray!8}Chronic kidney disease&24.78&27.96&24.05\\
Chronic obstructive pulmonary disease and bronchiectasis&17.33&20.41&15.84\\
\rowcolor{gray!8}Complications of surgical procedures or medical care&25.95&24.08&27.13\\
Conduction disorders&11.26&13.47&11.37\\
\rowcolor{gray!8}Congestive heart failure; nonhypertensive&33.41&34.69&32.09\\
Coronary atherosclerosis and other heart disease&31.15&33.47&33.14\\
\rowcolor{gray!8}Diabetes mellitus with complications&11.72&14.08&12.51\\
Diabetes mellitus without complication&20.46&19.80&21.85\\
\rowcolor{gray!8}Disorders of lipid metabolism&39.18&47.76&41.02\\
Essential hypertension&43.46&41.43&44.03\\
\rowcolor{gray!8}Fluid and electrolyte disorders&52.93&49.39&51.75\\
Gastrointestinal hemorrhage&7.51&7.96&7.23\\
\rowcolor{gray!8}Hypertension with complications and secondary hypertension&22.68&25.51&22.01\\
Other liver diseases&17.94&14.90&19.01\\
\rowcolor{gray!8}Other lower respiratory disease&13.62&14.49&14.95\\
Other upper respiratory disease&7.51&8.57&5.77\\
\rowcolor{gray!8}Pleurisy; pneumothorax; pulmonary collapse&11.86&12.86&10.64\\
Pneumonia&24.27&24.90&23.72\\
\rowcolor{gray!8}Respiratory failure; insufficiency; arrest&37.09&39.59&38.18\\
Septicemia&27.16&25.92&27.38\\
\rowcolor{gray!8}Shock&23.65&25.71&23.07\\
\bottomrule
\end{tabular}
}
\end{table}
\clearpage

\section{}
\label{app:checklist}
The concerns raised in this study have been summarized into a cohesive, condensed list of steps and questions to prevent the mentioned pitfalls and to design and validate preprocessing pipelines for MIMIC. The first section lists questions about the study design, on which some preprocessing choices depend. Not all of the points raised here might be applicable to every specific use-case or needed in every scenario:
\begin{enumerate}[leftmargin=*]
    \item Research Question \& Prediction Design:
    \begin{itemize}[nosep,leftmargin=*]
        \item Clinical task: What outcome or clinical phenomenon is being predicted or studied?
        \item Intended use: At what point in the clinical workflow would the prediction be made, by whom, and for what decision?
        \item Unit of analysis: Is the prediction made for a patient, hospital admission, ED stay, ICU stay, or individual time point?
        \item Clinical expertise: Do endpoint definition, variable selection, preprocessing, or evaluation require clinical input?
        \item Data perspective: Should the model represent the physiological patient state, the clinician’s knowledge, or both as separate representations?
        \item Prediction design: Are the prediction time, observation window, prediction gap, outcome window, and label-ascertainment time explicitly defined?
        \item Episode eligibility: How are patients with early outcomes, early discharge, transfers, or incomplete observation windows handled?
        \item Endpoint validity: Can the selected features and modalities plausibly evidence the endpoint at prediction time?
        \item Endpoint specificity: Does the endpoint combine clinically distinct phenomena that should instead be analyzed as separate labels or subgroups?
    \end{itemize}
    \item Cohort Construction \& Preprocessing:
    \begin{enumerate}[nosep,leftmargin=*]
        \item Cohort \& Episode Construction:
        \begin{itemize}[leftmargin=*]
            \item A care-pathway spine is constructed and validated across subject\_id, hadm\_id, and stay\_id.
            \item ED-to-hospital linkage, hospital admission, ICU admission, services, and transfers are checked for unmatched links, one-to-many mappings, overlaps, boundary-only events, and events outside the linked episode.
            \item Cohort attrition and coverage are reported separately for each module and modality.
            \item The release version of each MIMIC-IV module is recorded separately, and linkage and coverage are verified.
        \end{itemize}
        \item Temporal Alignment:
        \begin{itemize}[leftmargin=*]
            \item For every feature family, clinical event time, documentation time, and the best available proxy for clinician availability are determined.
            \item Timestamp selection matches the chosen modeling perspective.
            \item Event-to-availability delays are quantified where multiple timestamps exist.
            \item Point and interval boundary conventions are predefined.
            \item Ongoing events are censored at prediction time; future stop times, durations, or final statuses are not used.
            \item Features are included only if they were collected and available within the observation window according to the selected timestamp rule.
        \end{itemize}
        \item  Workflow-specific Representations:
        \begin{itemize}[leftmargin=*]
            \item When a clinical concept has multiple workflow representations, the selected table reflects the stage required by the modeling objective.
            \item For medications, orders, prescriptions, pharmacy processing, scheduled administrations, actual administrations, and documentation times remain separate.
            \item Agreement and disagreement between medication workflow tables are quantified.
            \item Current-admission diagnosis and procedure codes are treated as retrospective discharge information unless an independently valid availability time exists.
            \item Historical diagnosis and procedure codes are used only from episodes completed before the current prediction time.
        \end{itemize}
        \item Discharge-note Processing:
        \begin{itemize}[leftmargin=*]
            \item The temporal validity of each discharge-note section is assessed separately.
            \item A section is not assumed to contain admission-time knowledge solely because of its heading.
            \item Discharge notes are processed using an explicit heading inventory, order-aware extraction, and handling of optional, skipped, repeated, empty, malformed, and de-identified headings.
            \item Retained text spans are audited for retrospective additions, copy-forward content, extraction errors, and discharge-only information.
        \end{itemize}
        \item Data Partitioning:
        \begin{itemize}[leftmargin=*]
           \item All admissions, stays, and notes from the same patient remain in one train, validation, or test partition.
           \item Imputation, scaling, feature selection, vocabulary or heading discovery, threshold selection, and all other learned preprocessing are fitted using training data only.
        \end{itemize}
        \item  Missingness, Units,  \& Outliers:
        \begin{itemize}[leftmargin=*]
            \item Measurement units, item mappings, reference ranges, duplicate measurements, and implausible values are validated before aggregation or imputation.
            \item Missingness is quantified by feature, time window, module, and relevant clinical subgroup.
            \item Structural module-level missingness is distinguished from patient-level missingness.
            \item Imputation is not transferred from an observed module-specific subgroup to a population in which that modality was structurally unavailable.
            \item Imputation strategies are selected according to the feature and missingness mechanism, and fitted using training data only.
            \item Missingness indicators are considered, as missingness may contain clinically relevant information.
            \item Feature or stay exclusion thresholds are prespecified and justified; the prevalence of excluded items is reported.
        \end{itemize}
    \end{enumerate}
    \item Verification, Evaluation \& Documentation
    \begin{enumerate}[nosep,leftmargin=*]
        \item Model \& Modality Evaluation:
        \begin{itemize}[leftmargin=*]
            \item Unimodal baselines and modality-contribution methods are used to verify multimodal contribution.
            \item Unimodal and multimodal models are compared on identical evaluation episodes and data partitions.
            \item Where patients have multiple episodes, confidence intervals and statistical comparisons account for patient-level clustering.
            \item Endpoint subgroups are statistically examined and reviewed with clinical input where necessary.
        \end{itemize}
        \item Pipeline Auditing \& Reproducibility
        \begin{itemize}[leftmargin=*]
            \item Cohort counts and outcome prevalence are reported after every major inclusion, exclusion, and join step.
            \item Join attrition, duplicate amplification, unmatched identifiers, invalid timestamps, out-of-episode events, extraction failures, and exclusions are reported.
            \item Variable-selection criteria, aggregation rules, unit harmonization, outlier handling, modeling perspective, episode anchors, and prediction anchors are documented.
            \item Source tables, timestamp columns, availability rules, workflow stages, imputation strategies, and modality-coverage rules are documented for every feature family.
            \item Exact module versions, code, feature mappings, software versions, random seeds, and cohort definitions are provided.
            \item Preprocessing failure cases are reported quantitatively and, if suitable, through representative examples.
            \item Limitations and residual leakage, selection-bias, measurement-bias, and generalizability risks are explicitly acknowledged.
        \end{itemize}
    \end{enumerate}
\end{enumerate}

\end{document}